\documentclass[11pt,letterpaper]{article}
\usepackage[margin=1in]{geometry}
\usepackage[T1]{fontenc}
\usepackage[utf8]{inputenc}
\usepackage{mathptmx}
\usepackage[scaled=0.92]{helvet}
\usepackage{courier}
\usepackage{amsmath,amssymb}
\usepackage{booktabs}
\usepackage{array}
\usepackage{graphicx}
\usepackage{xcolor}
\usepackage{caption}
\usepackage{enumitem}
\setlist{nosep,leftmargin=*}
\usepackage[round]{natbib}
\usepackage[hyphens]{url}
\usepackage{hyperref}
\hypersetup{colorlinks=true,linkcolor=blue!50!black,urlcolor=blue!50!black,citecolor=blue!50!black}
\newcommand{\pendinginput}[2]{%
  \IfFileExists{#1}{\input{#1}}{%
    \par\noindent\fbox{\parbox{0.94\columnwidth}{\textbf{[PENDING]} #2}}}}

\newcommand{\appIdentify}{Appendix~\ref{app:identifying-change}}
\newcommand{\appConstruction}{Appendix~\ref{app:construction}}
\newcommand{\appCodebook}{Appendix~\ref{app:codebook}}
\newcommand{\appCorpus}{Appendix~\ref{app:capc-versions}}
\newcommand{\appTraining}{Appendix~\ref{app:training-bias}}
\newcommand{\appTechniques}{Appendix~\ref{app:techniques}}
\newcommand{\appPacket}{Appendix~\ref{app:adjudication}}
\newcommand{\appAI}{Appendix~\ref{app:aipolicy}}
\newcommand{\appContext}{Appendix~\ref{app:context}}
\newcommand{\appPrompts}{Appendix~\ref{app:prompts}}
\newcommand{\appData}{Appendix~\ref{app:data}}
\newcommand{\appDomainDir}{Appendix~\ref{app:domain-vs-direction}}
\newcommand{\appFormulation}{Appendix~\ref{app:twoaxis-note}}
\newcommand{\appColor}{Appendix~\ref{app:color-fusion}}

\title{Dual-Axis Policy Entailment: A Benchmark for Policy-Change Reasoning}
\author{Charles Chang\\Duke Kunshan University
\and Yuen Yuen Ang\\Johns Hopkins University
\and Bolun Sun\\Northwestern University}
\date{}

\begin{document}
\maketitle

\begin{abstract}
Whether a policy shifts domains and, by extension, continues, broadens,
narrows, or reverses previous policy is of paramount importance to the study
of public policy. We propose \emph{Dual-Axis Policy Entailment} (DAPE), a
``kill two birds with one stone'' method that evaluates an ordered pair of
policy directives (policy paragraphs in our corpus)---an earlier directive
$A$ and a later directive $B$---on both dimensions at once. A binary
judgment determines whether the two directives govern the same policy domain;
a five-way judgment determines whether $B$ continues, broadens, narrows, or
reverses the policy in $A$, or governs a different domain altogether, in
which case the relation is \emph{irrelevant}. Read together, the two judgments
form one five-category label---our primary metric---that identifies what the
later directive does to the earlier policy. The two-axis design does double
duty. Substantively, it captures two central dimensions of policy change:
domain shifts, which are common because governments must pivot, and the
direction of change. It does so without conflating errors on either dimension
into a generic ``irrelevant'' label. Methodologically, it makes annotation
more rigorous: a coder cannot dismiss a difficult pair as domain-irrelevant
without an explicit, checkable decision, removing an easy escape from the
directional judgment.
We introduce an expert-annotated benchmark of 1,000 ordered directive pairs
drawn from a large corpus of Chinese government policy, together with an
annotation-calibration protocol (an adjudication packet with pre/post
agreement measurement) that makes the task reliably codable and that we
contribute as a methodological artifact. Fourteen open-weight models and four
frontier API flagships fail the task zero-shot: even the best frontier
flagship's chance-corrected agreement with expert labels is far below the
human ceiling (Cohen's $\kappa \approx 0.29$ vs.\ $0.53$).
A fine-tuned open-weight model (LoRA SFT of Qwen3-14B, five-fold
cross-validation) closes this gap, reaching the coder-mutual human ceiling
($\kappa = 0.524$ vs.\ $0.527$).
We release all labels, splits, prompts, model outputs, and scoring scripts.
Finally, we apply our method to China's AI policy (1980--2024), one of the
world's most consequential AI-policy programs, and map the direction of policy
change across eras.
\end{abstract}

\section{Introduction}

Policy documents are usually analyzed one at a time, through close reading
\citep{freeman2011documents,dalglish2020document} or, more recently,
computational classification of topics, positions, and instruments
\citep{grimmer2013text,wilkerson2017large,sun2026capccg}. But policies are
issued to effect change, and many of the most substantively important
questions about them are \emph{relational}: does a later directive merely
continue an earlier program, extend its coverage, relax its obligations, or
even rescind it? Governments change policy frequently, yet such change is
hard to trace. Two documents may share formulaic administrative language
while regulating unrelated domains, and a reversal may be issued by a
different agency, under a different topic label, and in the vocabulary of an
entirely different policy area, obscuring the change and avoiding an explicit
admission of error. The measurement problem is well known in policy research:
observing the magnitude and direction of policy change is a problem of
conceptualization, not data alone \citep{greenpedersen2004dependent}. For
computational systems to identify such change rigorously, policy experts must
first define and exemplify what counts as real change.

\begin{figure*}[t]
\centering
\includegraphics[width=\textwidth]{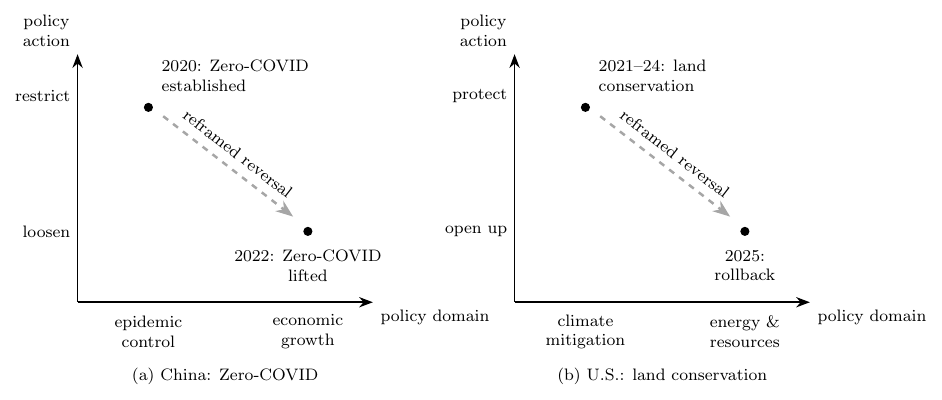}
\caption{Policy change decomposes into movement along two axes: a
\emph{policy-domain} axis (horizontal) and a \emph{policy-direction} axis
(vertical). (a)~China's Zero-COVID policy, established in 2020 as a
restriction in the domain of epidemic control \citep{nhc2020announcement}
and lifted in 2022 in the vocabulary of economic growth and social
stability \citep{jpcm2022newten}. (b)~U.S.\ land-conservation measures,
adopted as climate-change mitigation \citep{eo14008} and rolled back in the
name of energy security \citep{eo14154}. In both cases the reversal is a
diagonal move: the later directive shifts domain framing and action
direction at once. A single-axis model cannot say which axis moved; DAPE
scores them separately.}
\label{fig:shift}
\end{figure*}

Consider two reversals (Figure~\ref{fig:shift}). China's Zero-COVID policy
was, at its 2020 inception, unambiguously an \emph{epidemic-control} measure
\citep{nhc2020announcement}; when it was lifted in late 2022, the lifting was
justified not as a public-health reversal but in the language of
\emph{economic growth and social stability} \citep{jpcm2022newten}. The
pattern is not specific to China: U.S.\ land-conservation measures advanced
as \emph{climate-change mitigation} \citep{eo14008} were revoked in the name
of \emph{energy security and resource development} \citep{eo14154}. In both
cases the later directive differs from the earlier one in content, scope, and
wording while clearly targeting the same underlying policy domain. Yet the
two directives bear little linguistic similarity, making the relation
especially difficult to recover computationally.

These examples motivate the paper's central design: trace policy change not
on one dimension, but on two at once---\emph{do the two directives, however
dissimilar they appear, govern the same policy domain?}, and \emph{if so,
what does the later directive do to the earlier policy?}. We therefore
formulate \textbf{Dual-Axis Policy Entailment} (DAPE): an ordered directive
pair receives a binary policy-domain matching judgment and a five-way
policy-direction change judgment, with the constraint that the direction is
\emph{irrelevant} exactly when the domains differ. The collapsed five-category
label is our primary metric throughout, and we retain the axis separation as
a diagnostic frame. This separation prevents domain-matching failures from
being silently counted as direction failures: zero-shot models fail the two
axes in family-specific ways, and gating on the domain-match judgment alone
recovers roughly half of the observed error (Section~\ref{sec:analysis};
\appDomainDir{}).

We study this task in Chinese government policy, where consequential
directives emerge from a comparatively opaque process that combines frequent
central intervention with extensive local adaptation and implementation
discretion \citep{oi2018zouping,ang2023ambiguity}. China's AI-policy
trajectory is central to current technology-policy debates yet remains
largely unmapped at the directive level.

\paragraph{Contributions.} We make two contributions.
\begin{enumerate}
\item \textbf{A dual-axis, expert-annotated benchmark and calibration
protocol.} We construct 1,000 ordered pairs of Chinese policy directives,
expert-labeled on the two-axis task, with rare directional classes
(restriction, expansion, reversal) deliberately represented. The calibration
exercise---an adjudication process with a training guide, worked examples,
and blind pre/post agreement measurement---reveals what can and cannot be
measured. To our knowledge, DAPE is the first benchmark for formally measuring
directive-level policy change, enabling applications such as tracking policy
age and studying the policymaking process.
\item \textbf{An expert-assisted training protocol that reaches the human
ceiling.} Starting from the expert labels, we train a LoRA-adapted open-weight
model (Qwen3-14B) with document-group-disjoint cross-validation. The resulting
classifier matches the coder-mutual human ceiling on chance-corrected
agreement, while every zero-shot system---including GPT-5, Gemini 2.5 Pro,
Claude Sonnet 4.5, and DeepSeek-V3---falls well short of it on both axes.
\end{enumerate}

\paragraph{Why this matters.} Automated direction-of-change labels enable
three measurements that are impractical by hand: detecting quiet rollbacks
pursued through low-salience revisions
\citep{weaver1986politics,pierson1994dismantling,hacker2004privatizing};
monitoring where local reissues narrow central policy \citep{oi2018zouping};
and supplying the timing and direction data required for credible program
evaluation. China's AI policy is our motivating domain (Section~7; \appAI{}).

\paragraph{Headline findings.} (i) The five-category task is reliably codable
by trained experts after calibration (coder-mutual $\kappa = 0.527$).
(ii) Zero-shot LLMs, open or frontier, do not solve it (best zero-shot
$\kappa \approx 0.34$; frontier flagships $\kappa \leq 0.29$).
(iii) Expert-assisted fine-tuning of an open-weight 14B model reaches the
human ceiling ($\kappa = 0.524$ vs.\ $0.527$) and beats the best zero-shot
model by $+0.18$ macro-F1.

\section{Related Work}
\paragraph{NLI and document-level inference.}
Natural language inference established entailment, contradiction, and
neutrality as general-purpose semantic relations
\citep{dagan2006pascal,bowman2015large,williams2018broad}. Inference becomes
harder when evidence is distributed across long, specialized texts
\citep{koreeda2021contractnli,chalkidis2022lexglue}, although research on
more formally structured legal documents has shown promise
\citep{rabelo2022overview,goebel2025competition}. Our task shares NLI's
ordered-pair form, but uses policy-directive pairs and separates domain
matching from the \emph{direction of policy change}.

\paragraph{Legal and regulatory change detection.}
Closest to our task is work that classifies how legal texts change:
RegTrack \citep{yu2026regtrack} introduces a fine-grained benchmark for
multi-class legal change detection, and legal NLP has long studied retrieval
and entailment over statutes and case law
\citep{chalkidis2020legal,zhong2020does,kim2015overview,goebel2024overview}.
We differ in three ways. First, our \emph{administrative policy directives}
are shorter-lived, heavily reissued, and routinely paraphrased; they can also
change more abruptly than statutes. Second, our labels encode the
\emph{direction} of change that experts can reliably code, not textual-edit
categories. Third, policy issuance in our non-Western authoritarian context
can be obscure and unpredictable: landmark shifts such as the late-1970s
economic reforms and the Zero-COVID reversal surprised many experts, making
change difficult to detect and anticipate.

\paragraph{Policy change, diffusion, and measurement in political science.}
The study of policy change spans incrementalism
\citep{lindblom1959muddling}, punctuated agendas
\citep{baumgartner1993agendas,kingdon1984agendas}, advocacy coalitions
\citep{sabatier1988advocacy}, policy paradigms \citep{hall1993policy}, and
retrenchment through blame avoidance and drift
\citep{pierson1994dismantling,weaver1986politics,hacker2004privatizing,pierson2000increasing}.
Text-as-data methods measure positions and topics
\citep{grimmer2013text,laver2003extracting,slapin2008scaling,wilkerson2017large},
and work on China documents adaptive implementation and local
experimentation
\citep{heilmann2011mao,ang2016china,ang2023ambiguity,oi2018zouping}.
These literatures theorize policy change but lack directive-level,
direction-labeled data; our benchmark supplies exactly that measurement layer
for one national system.

\paragraph{LLMs as annotators.}
LLMs can serve as effective annotators for social-science text, but
performance depends on task definition, label balance, and validation against
human judgment \citep{gilardi2023chatgpt,ziems2024large}. Our results sharpen
this: zero-shot annotation fails on directional policy change even for
frontier models, while expert-assisted fine-tuning succeeds---arguing for
pipelines in which expert labels calibrate open models rather than being
replaced by API calls.

\section{Task: Dual-Axis Policy Entailment}
Each instance is an ordered pair $(A, B)$ of \emph{policy directives}: policy
paragraphs, each carrying a distinctive directive within a possibly longer
document, together with the source document's metadata (title, issuing
agency, year, topic fields) and the paragraph text. $B$ is issued
after $A$. The model produces two outputs: a binary \emph{policy-domain
matching} judgment $d \in \{0,1\}$---do the two directives govern the same
underlying program, regime, or initiative, however the domain is named or
framed?---and a \emph{policy-direction change} judgment $r$ describing what
$B$ does to the policy established in $A$:

\begin{itemize}
\item \textbf{Irrelevant} (stored label \textsc{unrelated}): $B$ concerns a
different policy domain. Shared formulaic language or topical overlap alone
does not count as related.
\item \textbf{Consistency} (\textsc{consistency}): $B$ continues the same
policy---a renewal or parallel reissue with no material change.
\item \textbf{Restriction} (\textsc{restriction}): $B$ narrows the
policy---reduced coverage, relaxed obligations, or lower stringency.
\item \textbf{Expansion} (\textsc{expansion}): $B$ broadens the
policy---wider coverage, geographic reach, resources, obligations, or
stringency.
\item \textbf{Reversal} (\textsc{contradiction}): $B$ rescinds or materially
reverses the earlier policy.
\end{itemize}

\begin{table}[t]
\centering
\footnotesize
\setlength{\tabcolsep}{3pt}
\begin{tabular}{p{0.20\columnwidth} p{0.55\columnwidth} r}
\toprule
\textbf{Label} & \textbf{What $B$ does to the policy in $A$} & \textbf{$n$} \\
\midrule
Consistency & Continues the same policy: renewal or parallel reissue, no material change. & 255 \\
Expansion & Broadens coverage, reach, resources, obligations, or stringency. & 171 \\
Restriction & Narrows coverage, relaxes obligations, lowers stringency. & 176 \\
Reversal & Rescinds or materially reverses the earlier policy. & 127 \\
Irrelevant & A different policy domain; formulaic or topical overlap alone does not count. & 271 \\
\bottomrule
\end{tabular}
\caption{The five policy-change labels and their counts in the 1,000-pair
benchmark. The five rows form the policy-direction change axis; the binary
policy-domain matching axis's single mismatch outcome is definitionally the
\emph{irrelevant} row ($d = 0 \Leftrightarrow$ \textsc{unrelated}), so both
axes are represented. Prose terms are used throughout; the stored labels are
\textsc{consistency}, \textsc{expansion}, \textsc{restriction},
\textsc{contradiction} (reversal), and \textsc{unrelated} (irrelevant).
}
\label{tab:labels}
\end{table}

The outputs are constrained: the relation is \emph{irrelevant} exactly when
the domains differ ($d = 0 \Leftrightarrow r = \textsc{unrelated}$). Reading
the two outputs together therefore yields one label per pair---\emph{irrelevant}
when $d = 0$, otherwise the directional judgment $r$---which we call the
\emph{collapsed five-category label}. It is our primary label and metric: it
states what the later directive does to the earlier policy, and it is directly
comparable to single-task formulations. The binary axis $d$ remains
available as a diagnostic: scoring the two axes separately reveals where
models actually fail, and we show that zero-shot failures on the domain axis
are family-specific and large enough to masquerade as direction errors under
a collapsed reading (\appDomainDir{}).

The separation is also the better \emph{annotation} instrument. Under a
single five-way scheme, a coder facing a pair like those in
Figure~\ref{fig:shift}---same underlying policy, reframed in another
domain's vocabulary---cannot label it a reversal without close reading,
while \emph{irrelevant} takes only a glance at the divergent topics---so a
collapsed scheme prices the two errors asymmetrically and absorbs
boundary-crossing reversals into \emph{irrelevant}. DAPE removes that
shortcut:
the domain question is asked first and explicitly, so \emph{irrelevant}
requires an affirmative decision that the domains differ, and the
directional judgment is requested only for pairs that survive that gate
(models suffer the same conflation; \appDomainDir{}).

The label set is deliberately small and directional. Reversal is reserved for
material rescission or rollback, and is the only class with a reliable
textual cue (repeal or cessation language citing a prior document
identifier); restriction and expansion require comparing the \emph{degree} of
regulation rather than spotting a keyword, which makes them the hardest
classes for both humans and models. Because consistency and irrelevance
dominate the natural distribution of directive pairs, raw accuracy is
uninformative; we report macro-F1 and chance-corrected Cohen's $\kappa$ as
the primary metrics throughout. \appIdentify{} develops the identification
problem in detail, including how domain shifts and deliberately understated
wording can hide consequential changes from public view. The two outputs can
be elicited jointly, in cascade, or as a single direct judgment;
\appFormulation{} finds the three strategies predictively comparable, which
is why we score the collapsed relation as the primary metric while retaining
the axis separation for diagnosis.

\section{The Benchmark}
\subsection{Corpus and Candidate Generation}
The benchmark is drawn from the Chinese Adaptive Policy Corpus (CAPC)
\citep{sun2026capccg}. Because the corpus exists in several versions with
very different coverage, we state the corpus definitions explicitly. The published
CAPC release (CAPC-CG) covers central party and government directives
(1949--2023). Our pair-mining frame is an extended central-only snapshot that
adds ministerial document types through 2024. A separate, ongoing collection
effort integrates a leading commercial policy database with documents
scraped from government websites across
national, ministerial, provincial, prefectural, and county levels; after
verified title-overlap removal, its core contains 4,997,123 documents.
\appCorpus{} reports the full component and coverage comparison.
Candidate pairs are formed from BGE document embeddings
\citep{xiao2024cpack}, stratified across cosine-similarity bands, agency
match, and year gap, so that both hard related pairs (low-similarity genuine
reissues) and hard unrelated pairs (high-similarity administrative templates)
are represented. Because directional change is rare in the natural
distribution, later sampling rounds deliberately oversampled restriction,
expansion, and especially reversal (repeal, lapse, or non-enforcement
language co-occurring with a cited prior-document identifier, or
\emph{wenhao}). The benchmark therefore evaluates classification
\emph{conditional on candidate generation}; it is not an estimate of the
corpus-wide prevalence of policy relations. The trained classifier's behavior
on an unsupplemented frame of newly collected documents is reported in
\appCorpus{}. \appConstruction{} details the four sampling rounds, the
rare-class oversample, and the document-group-disjoint split discipline,
together with natural-pair and alternate-embedder sensitivity analyses.
Existing early-round and strict-recode labels support the human-referenced
comparisons; newly constructed natural pairs are used only for composition
analysis, not for unvalidated accuracy claims.

\subsection{Annotation Protocol and Calibration Exercise}
Every retained pair was labeled by expert annotators trained in policy
analysis, following a written codebook whose central gate is a
\emph{shared policy domain}: two directives are related only if they regulate
the same program, regime, or initiative; same topic, different region or
hierarchy level, or parallel sibling implementations of one parent policy are
irrelevant by rule. Directional labels are assigned on a single-label ladder
with a conservative-shift rule for mixed changes. \appCodebook{} reproduces
the full five-class codebook with worked examples.

A persistent difficulty in policy-change annotation is that independent
coders apply the relevance boundary differently. We therefore treat
\emph{calibration} as a first-class part of the protocol and contribute it as
a methodological artifact. After an initial reliability study (below)
revealed systematic divergence at the relevance boundary, we built an
\emph{adjudication packet}: a training guide stating the shared-domain gate,
the reversal cue list, and mixed-change conventions; a 90-pair training set
with both coders' original labels, the gold label, and written rationales;
and a 30-pair blind calibration test on which agreement is measured
\emph{before and after} training. The packet operationalizes adjudication as
a measurable procedure---agreement is re-measured on held-out pairs after
calibration, against both the other coder and the held-out gold---rather than
as an unrecorded discussion. \appPacket{} describes the packet in full.
On the returned blind tests, two newly trained coders converge at five-way
$\kappa = 0.519$ (binary $\kappa = 0.667$) on a deliberately boundary-loaded
test (\appPacket{}).

\subsection{Inter-Annotator Agreement, Reported Honestly}
A second expert independently re-labeled a blind, stratified 120-pair subset
under the finalized codebook (one pair dropped as incomplete, $n=119$).
The two coders agree on the binary related/irrelevant judgment at 88.2\%
($\kappa = 0.730$) and on the five-way label at 75.6\% ($\kappa = 0.527$);
within the graded classes (consistency, restriction, expansion) they agree on
the specific direction in 7 of 20 cases where both assigned a graded label.
We also report Gwet's AC1 alongside $\kappa$ because $\kappa$ is unstable
under extreme label marginals \citep{gwet2008computing}.

Two findings from the reliability study shape how the benchmark should be
read. First, \emph{the coders are stricter than the gold standard at the
relevance boundary}: against the first-pass gold, each coder's binary
agreement is only $\sim$43--45\%, because the gold was annotated under an
earlier, deliberately liberal instruction (``mark unrelated only for clearly
different subjects'') while the finalized codebook requires a shared policy
domain; both blind coders independently tightened the boundary (marking only
36/119 and 40/119 pairs related) and converged on each other while doing so.
The codebook, not individual coder taste, resolves this divergence. A further
173 pairs have been recoded under the strict rule with per-pair rationales.
We use the nonblank recodes as a strict-rule sensitivity reference, while 26
missing or cross-round-conflicted pairs remain flagged rather than being
assigned unvalidated labels. The single-coder recodes are not replacement gold
(\appPacket{}). Second, \emph{graded directional change is intrinsically hard}: reversal is
the most stable class in both coder-vs-coder and coder-vs-gold comparisons
(consistent with its explicit textual cue), while restriction and expansion
resist reliable coding even by experts. We therefore report model performance
separately on the pairs where the two coders agree and disagree, and treat
the coder-mutual $\kappa$ as the operative human ceiling.

\section{Experiments}
\subsection{Setup}
All models receive the same task inputs: the source document's title, agency,
year, and topic metadata, plus the first 1,500 characters of each directive's
text. Two output formats are used: the direct collapsed
five-category judgment (all headline runs) and the joint two-output format
(binary domain match plus relation) for the dual-axis arms; both are scored
on the collapsed relation for comparability, and the dual-axis runs are
additionally scored on each axis separately.
Decoding is greedy unless stated; unparsed or refused
outputs count as errors so that every system is scored on every pair (we
additionally report structured-output coverage). We report macro-F1 and
Cohen's $\kappa$ on all 1,000 gold pairs; the class distribution is
restriction/expansion/reversal-enriched relative to the natural pair
distribution, so absolute scores are conditional on the enriched frame while
system rankings can vary across evaluation subsets. All prompts are listed
verbatim in \appPrompts.

\paragraph{Baselines.} A majority predictor (always irrelevant) and BGE-m3
cosine-similarity thresholding \citep{chen2024bgem3} anchor the low end.

\subsection{Zero-Shot and Few-Shot Results}
\begin{table*}[t]
\centering
\footnotesize
\setlength{\tabcolsep}{4pt}
\caption{Five-category policy-change classification on the 1,000-pair gold
benchmark: open-weight and API models under the main prompt, with formulation
variants and baselines. Coverage is the share of pairs with a parseable
prediction; rows below full coverage are scored on parsed pairs and are not
directly comparable. Macro-F1 and Cohen's $\kappa$ are the primary metrics.
Script-generated; do not hand-edit.}
\label{tab:zeroshot}
\begin{tabular}{l l r r r r}
\toprule
\textbf{Model} & \textbf{Tier} & \textbf{Cov\%} & \textbf{Acc} & \textbf{Macro-F1} & \textbf{Cohen's $\kappa$} \\
\midrule
qwen3-32b & open & 100 & 0.480 & 0.452 & 0.337 \\
gemini-2.5-flash & API & 100 & 0.452 & 0.430 & 0.297 \\
gpt-5-mini & API & 100 & 0.425 & 0.383 & 0.279 \\
qwen2.5-14b-instruct & open & 100 & 0.412 & 0.357 & 0.266 \\
qwen3-8b & open & 100 & 0.385 & 0.353 & 0.214 \\
llama-3.3-70b & API & 100 & 0.403 & 0.348 & 0.230 \\
qwen3-14b & open & 100 & 0.388 & 0.342 & 0.202 \\
deepseek-v3 & API & 100 & 0.393 & 0.327 & 0.210 \\
mimo-7b-rl & open & 92 & 0.383 & 0.306 & 0.208 \\
deepseek-r1-distill-qwen-14b & open & 100 & 0.366 & 0.301 & 0.159 \\
glm-4-9b-chat & open & 100 & 0.326 & 0.290 & 0.245 \\
gemma-2-9b-it & open & 100 & 0.345 & 0.271 & 0.163 \\
qwen3-4b & open & 100 & 0.323 & 0.238 & 0.106 \\
qwen2.5-7b-instruct & open & 100 & 0.344 & 0.230 & 0.126 \\
internlm2\_5-7b-chat & open & 100 & 0.344 & 0.184 & 0.111 \\
mistral-7b-instruct-v0.2 & open & 84 & 0.226 & 0.180 & 0.182 \\
BGE-m3 cosine (t=0.78) & base & 87 & 0.351 & 0.180 & 0.092 \\
gemma-7b-it & open & 100 & 0.330 & 0.173 & 0.092 \\
llama-3.1-8b-instruct & open & 100 & 0.269 & 0.115 & 0.021 \\
Majority (always irrelevant) & base & 100 & 0.271 & 0.085 & 0.000 \\
\bottomrule
\end{tabular}

\end{table*}

Table~\ref{tab:zeroshot} reports the open-weight lineup. The best zero-shot
system, Qwen3-32B, reaches macro-F1 0.452 and $\kappa = 0.337$---well above
the majority baseline ($\kappa = 0$) but at only about $0.64\times$ the
coder-mutual human ceiling ($\kappa = 0.527$).
Smaller open models degrade steeply, and the embedding baseline barely
exceeds chance on the directional classes.

\begin{table*}
\centering
\scriptsize
\setlength{\tabcolsep}{4pt}
\caption{Frontier API flagships under the direct five-category prompt (1,000 gold pairs, canonical 1,500-character window). Few-shot, multi-seed, and context-length variants are in Appendix~\ref{app:context}.}
\label{tab:waveb}
\resizebox{\textwidth}{!}{%
\begin{tabular}{lrrrrrrrrrr}
\toprule
slug & n\_scored & coverage & macroF1 & kappa & acc & F1\_irrelevant & F1\_consistency & F1\_restriction & F1\_expansion & F1\_reversal \\
\midrule
claude\_sonnet\_45\_\_direct5\_\_t1500 & 1000 & 0.999 & 0.412 & 0.265 & 0.432 & 0.511 & 0.411 & 0.217 & 0.446 & 0.474 \\
deepseek\_chat\_\_direct5\_\_t1500 & 1000 & 1.000 & 0.406 & 0.282 & 0.440 & 0.527 & 0.400 & 0.124 & 0.477 & 0.503 \\
gemini\_25\_pro\_\_direct5\_\_t1500 & 1000 & 0.998 & 0.418 & 0.273 & 0.436 & 0.508 & 0.382 & 0.195 & 0.465 & 0.542 \\
gpt\_5\_\_direct5\_\_t1500 & 1000 & 1.000 & 0.420 & 0.256 & 0.417 & 0.495 & 0.200 & 0.357 & 0.433 & 0.618 \\
\bottomrule
\end{tabular}%
}
\end{table*}

Table~\ref{tab:waveb} reports frontier API flagships under the same direct
five-category prompt. All four fail zero-shot:
GPT-5, Gemini 2.5 Pro, Claude Sonnet 4.5, and DeepSeek-V3 all land at
macro-F1 $0.39$--$0.42$ and $\kappa \approx 0.26$--$0.28$.
Variant arms---a class-balanced five-shot condition with a corrected class
prior, a multi-seed robustness arm (temperature 0.7, two seeds), and a
context-length sweep---change none of this: five-shot prompting helps at
most modestly (best: GPT-5, macro-F1 0.445), seed-to-seed spread is $\leq
0.003$ macro-F1, and longer windows are flat (full variant table in
\appContext{}).

\IfFileExists{generated/waveM_scores.tex}{%
We also re-run the four flagships under the joint
two-output (dual-axis) prompt, scoring the binary domain axis and the
collapsed relation separately (full table in \appFormulation{}). Two findings
emerge. First, the elicitation
format itself matters little: collapsed five-way performance under the joint
prompt is within noise of the direct prompt for GPT-5 ($-0.004$ macro-F1)
and modestly lower for the others ($-0.03$ to $-0.08$), with the model
ranking preserved. Second, the joint prompt exposes the domain axis---and
with it, a measurable consequence of an erroneous relatedness prior:
the original prompt's asserted ``$\sim$90\% of pairs share a
topic'' pushes all four flagships to declare 95\%+ of pairs related,
compressing binary domain-match agreement to $\kappa = 0.13$--$0.17$ despite
$\sim$75\% accuracy, barely above the all-related baseline; a corrected 73\%
prior partially repairs the domain axis ($\kappa$ up on all four models)
while leaving the collapsed metric essentially unchanged (\appFormulation{}).
The axis
separation makes this failure visible; a collapsed-only reading would absorb
it into direction errors.
}{%
\par\noindent\fbox{\parbox{0.94\columnwidth}{\textbf{[PENDING]} Dual-axis
frontier results (waveM) --- pending the API runs.}}}

\subsection{Expert-Assisted Fine-Tuning}
We fine-tune Qwen3-14B \citep{yang2025qwen3} with LoRA
\citep{hu2022lora} on the expert-labeled pairs under stratified five-fold
cross-validation (80/20 folds), with folds made document-group-disjoint so
that no source document appears on both sides of a split. We compare the
plain SFT arm against two variants: SFT initialized from a domain-adaptive
pretrained checkpoint (DAPT+SFT), and SFT with model-labeled silver expansion
from a 34,871-pair unlabeled pool (gold + silver).

\begin{table*}[t]
\centering
\footnotesize
\setlength{\tabcolsep}{4pt}
\caption{Five-fold cross-validated fine-tuning results on the 1,000-pair expert gold set.}
\label{tab:wavec}
\resizebox{\textwidth}{!}{%
\begin{tabular}{lrrrrrrrrr}
\toprule
Model & $n$ & macro-F1 & $\kappa$ & acc. & F1 irrelevant & F1 consistency & F1 restriction & F1 expansion & F1 reversal \\
\midrule
SFT Qwen3-14B LoRA & 1000 & 0.634 & 0.524 & 0.626 & 0.586 & 0.710 & 0.544 & 0.551 & 0.780 \\
DAPT + SFT & 1000 & 0.621 & 0.509 & 0.614 & 0.561 & 0.698 & 0.548 & 0.513 & 0.786 \\
Gold + model-silver SFT & 1000 & 0.638 & 0.529 & 0.631 & 0.591 & 0.709 & 0.550 & 0.552 & 0.790 \\
\bottomrule
\end{tabular}%
}
\end{table*}

\IfFileExists{generated/waveN_eval_table.tex}{%
\begin{table}[t]
\centering
\footnotesize
\setlength{\tabcolsep}{3pt}
\caption{Five-fold cross-validated fine-tuning with the collapsed five-way
target versus the dual-axis (two-output) target, scored on the collapsed
relation. Script-generated; do not hand-edit.}
\label{tab:waven}
\resizebox{\columnwidth}{!}{%
\begin{tabular}{lrrrrrrrrr}
\toprule
Model & $n$ & macro-F1 & $\kappa$ & acc. & F1 irrelevant & F1 consistency & F1 restriction & F1 expansion & F1 reversal \\
\midrule
SFT Qwen3-14B LoRA (five-way) & 1000 & 0.634 & 0.524 & 0.626 & 0.586 & 0.710 & 0.544 & 0.551 & 0.780 \\
SFT Qwen3-14B LoRA (dual-axis, collapsed) & 1000 & 0.649 & 0.548 & 0.643 & 0.600 & 0.724 & 0.594 & 0.551 & 0.776 \\
\bottomrule
\end{tabular}
}
\end{table}
}{%
\par\noindent\fbox{\parbox{0.94\columnwidth}{\textbf{[PENDING]} Dual-output
SFT table (waveN) --- pending.}}}

\IfFileExists{generated/waveN_eval_table.tex}{%
Fine-tuning with the joint two-output target tests whether the axis
separation itself carries signal. It does: the dual-output variant reaches
macro-F1 0.649 and $\kappa = 0.548$ on the collapsed relation, slightly
\emph{above} the five-way recipe (0.634/0.524), and the explicit domain-match
head attains binary $\kappa = 0.467$ (vs.\ 0.431 for the implicit boundary in
the collapsed model) with perfect constraint consistency---in all 1,000
held-out predictions, the relation is irrelevant exactly when the domains
differ. The domain axis, in other words, is learnable as an auxiliary
supervision signal rather than a post-hoc constraint.
}{}

Table~\ref{tab:wavec} shows that fine-tuning makes the task tractable:
pooled five-fold SFT reaches macro-F1 0.634 and $\kappa = 0.524$, numerically
matching the coder-mutual human ceiling ($\kappa = 0.527$) and exceeding the
best zero-shot system by $+0.18$ macro-F1.
Domain-adaptive pretraining does not help further, and a model-labeled
silver expansion (2,000 pairs labeled by the fold-0 SFT model, not
additional expert annotation) yields a change within fold-to-fold noise. An
auxiliary-signal test using the corpus's paragraph-level policy-obligation
coding as side information for the direction axis likewise yields no gain,
and degrades the model when imposed as a hard rule (\appColor{}).
\appTraining{} complements this comparison with a direct zero-shot /
few-shot / fine-tuning regime comparison, an expert-annotation learning
curve, and label-bias analyses across model families, including whether
expert fine-tuning corrects family-specific biases. \appTechniques{}
compares preference-style objectives, rejection sampling, and test-time
self-consistency against the same recipe.

\paragraph{Jurisdiction-corrected gold.} Adjudication of 158 gold pairs
flagged by a jurisdiction audit flipped 60 labels to irrelevant.
Retrained on the corrected labels, the five-way and dual-axis recipes reach
$\kappa = 0.529$ and $0.540$ (domain-axis $\kappa = 0.536$, zero constraint
violations), so the dips seen when the original models were merely re-scored
were training-label artifacts (\appPacket{}).

\subsection{Human-Ceiling and Agreement-Subset Analysis}
Because label reliability bounds achievable agreement, we evaluate systems
relative to the coder-mutual ceiling rather than to an assumed perfect gold.
On the pairs where the two independent coders mutually agree, the best models
score roughly $2.4\times$ higher macro-F1 than on the pairs where the coders
disagree; model errors concentrate exactly where human judgment splits.
The fine-tuned classifier's pooled $\kappa$ of 0.524 sits at 99\% of the
coder-mutual ceiling (0.527), meaning residual model error is at the level of
inherent label ambiguity under the current codebook.

\subsection{Context-Length Curve}
Truncating each directive to 1,500 characters could in principle drop
decisive evidence. Two analyses test this. First, an evidence-coverage study
on full source texts locates the decisive repeal cue: it appears within the first
1,500 characters in 99\% of reversal pairs (median position 268 characters).
Second, extending the input window to 4,000 and 8,000 characters (and to full
text for DeepSeek-V3) leaves five-category performance flat or slightly
negative (Table~\ref{tab:waveb}, context-sweep rows; \appContext{}):
truncation is not the bottleneck for this task.

\section{Analysis}\label{sec:analysis}
\paragraph{Where the errors are: the domain axis vs.\ the direction axis.}
Decomposing each system's predictions into the two axes (\appDomainDir{})
shows zero-shot models failing
the domain boundary in family-specific ways: the Qwen models are
over-inclusive, keeping $\sim$98\% of gold-related pairs as related and
flooding the unrelated class, while GPT-5 is over-strict, recalling only
70.4\%. Once a pair is correctly held in the related set, zero-shot direction
accuracy (0.49--0.61) is markedly better than the collapsed scores suggest.
An oracle diagnostic makes the split quantitative: gating zero-shot
predictions on the gold domain-match judgment raises Qwen3-32B to macro-F1 0.618
($\kappa = 0.576$), so roughly half of the observed error sits on the domain
axis rather than in directional reasoning. Fine-tuning repairs both axes
(domain-match $\kappa$ 0.161 $\to$ 0.431; conditional direction accuracy
0.490 $\to$ 0.759) but leaves the boundary the harder judgment---binary
$\kappa = 0.431$ against the human boundary ceiling of $0.730$---mirroring
the annotation data, where the relevance boundary was the main source of
inter-coder divergence. A collapsed, single-axis reading would misattribute
every one of these boundary failures to direction.

\paragraph{Per-class difficulty.}
Restriction is the hardest class for every zero-shot system (per-class F1
$0.12$--$0.36$ across frontier models), followed by expansion; reversal is
the easiest change class zero-shot because of its explicit repeal cue.
Under SFT the same ordering holds, but the gap widens: reversal becomes
easier still (F1 0.780) while restriction/expansion remain hardest
(0.544/0.551), mirroring
the human data (reversal most reliably coded; graded direction least, 7/20).

\paragraph{Label-noise ceiling.}
The coder-mutual $\kappa$ of 0.527 provides the operative human ceiling for
agreement with single-annotator gold labels. The fine-tuned model reaches
this ceiling while zero-shot models plateau near half of it. This suggests
that further gains depend at least as much on adjudicated labels (in progress
via the calibration packet) as on larger models.

\section{Application: AI-Policy Change in China}
Finally, we apply the trained classifier to AI-related policy-directive pairs
in the integrated multi-level corpus (1980--2024; \appAI{}). Three findings
emerge. First, China's AI-policy trajectory is not the monotonic tightening
often portrayed in Western media: broadenings and narrowings interleave in
every era with sufficient data (Hu era, 36 expansions against 10
restrictions; Xi era, 1,147 against 86), expansion remains the dominant mode
of directional change even after 2012, and the 18 outright reversals appear
only in the Xi era. Second, directional change is concentrated at the
center---9.7\% of national and 7.6\% of ministerial pairs, against 1.6\% of
provincial pairs, whose dominant mode is continuity and implementation.
Third, provinces differ sharply: directional shares range from 15.5\%
(Anhui) and 8.4\% (Beijing) down to 0.5\% (Jilin), a thirty-fold spread that
national aggregates obscure (\appAI{} reports era, per-level, and
per-province breakdowns and a spot-checked sample; shares are
frame-conditional, with comparable ensemble confidence across frames).
On 20,000 pairs from government-website documents postdating the benchmark
(90\% from 2024 or later), the classifier recovers directional changes at
channel-expected rates, with eight of ten spot-checks correct (\appCorpus{}).

\section{Ethical Considerations and Limitations}
\paragraph{Limitations.}
First, the benchmark covers one national policy system; transfer to other
states and to statute-dominated corpora is untested (a binary-only COLIEE
check \citep{rabelo2022overview,goebel2025competition} suggests the schema
transfers to binary entailment but not necessarily the directional classes).
Second, the mined-pair frame oversamples hard and rare pairs relative to the
natural distribution, so absolute scores are frame-conditional and rankings
can also vary; newly collected documents (\appCorpus{}), the initial
less-enriched workbook, and a uniform natural-pair composition analysis
(\appConstruction{}) probe this sensitivity without using model-generated
reference labels for accuracy.
Third, inter-annotator agreement on the graded classes is moderate
($\kappa = 0.527$ mutual; 7/20 on the graded three-way split), and the
first-pass gold embodies a more permissive relevance boundary than the
finalized codebook; we report this openly, evaluate on agreement subsets,
and release an adjudication packet for recalibration.

\paragraph{Ethics.}
The corpus consists of publicly promulgated administrative and regulatory
documents, excluding classified and non-public intra-Party material by
construction. Model outputs are research measurements, not authoritative
legal or administrative interpretations, and automated change detection in
an authoritarian context could be misused; the release documents sampling
and annotation biases to support critical use.

\section{Conclusion}
DAPE separates policy-domain matching from policy-direction change, collapsed
into a five-category primary metric. Our 1,000-pair expert benchmark and
calibration protocol make inter-coder divergence measurable and correctable.
Zero-shot LLMs fail on both axes; expert-assisted fine-tuning reaches the
human ceiling. The residual difficulty lies at the domain boundary, where
human judgment also diverges. We release all labels, splits, prompts, model
outputs, and scoring scripts.

\paragraph{Data and code availability.}
Labels, splits, prompts, model outputs, and scoring scripts will be released
publicly (\appData{}).

\bibliographystyle{plainnat}
\bibliography{references}

\begin{thebibliography}{44}
\providecommand{\natexlab}[1]{#1}
\providecommand{\url}[1]{\texttt{#1}}
\expandafter\ifx\csname urlstyle\endcsname\relax
  \providecommand{\doi}[1]{doi: #1}\else
  \providecommand{\doi}{doi: \begingroup \urlstyle{rm}\Url}\fi

\bibitem[Ang(2016)]{ang2016china}
Yuen~Yuen Ang.
\newblock \emph{How China Escaped the Poverty Trap}.
\newblock Cornell University Press, Ithaca, 2016.

\bibitem[Ang(2024)]{ang2023ambiguity}
Yuen~Yuen Ang.
\newblock Ambiguity and clarity in {C}hina's adaptive policy communication.
\newblock \emph{The China Quarterly}, 257:\penalty0 20--37, 2024.
\newblock \doi{10.1017/S0305741023000826}.

\bibitem[Baumgartner and Jones(1993)]{baumgartner1993agendas}
Frank~R. Baumgartner and Bryan~D. Jones.
\newblock \emph{Agendas and Instability in American Politics}.
\newblock University of Chicago Press, Chicago, 1993.

\bibitem[Bowman et~al.(2015)Bowman, Angeli, Potts, and
  Manning]{bowman2015large}
Samuel~R. Bowman, Gabor Angeli, Christopher Potts, and Christopher~D. Manning.
\newblock A large annotated corpus for learning natural language inference.
\newblock In \emph{Proceedings of the 2015 Conference on Empirical Methods in
  Natural Language Processing (EMNLP)}, pages 632--642, Lisbon, Portugal, 2015.
  Association for Computational Linguistics.
\newblock \doi{10.18653/v1/D15-1075}.
\newblock URL \url{https://aclanthology.org/D15-1075/}.

\bibitem[Chalkidis et~al.(2020)Chalkidis, Fergadiotis, Malakasiotis, Aletras,
  and Androutsopoulos]{chalkidis2020legal}
Ilias Chalkidis, Manos Fergadiotis, Prodromos Malakasiotis, Nikolaos Aletras,
  and Ion Androutsopoulos.
\newblock {LEGAL-BERT}: The muppets straight out of law school.
\newblock In \emph{Findings of the Association for Computational Linguistics:
  EMNLP 2020}, pages 2898--2904. Association for Computational Linguistics,
  2020.
\newblock \doi{10.18653/v1/2020.findings-emnlp.261}.
\newblock URL \url{https://aclanthology.org/2020.findings-emnlp.261/}.

\bibitem[Chalkidis et~al.(2022)Chalkidis, Jana, Hartung, Bommarito,
  Androutsopoulos, Katz, and Aletras]{chalkidis2022lexglue}
Ilias Chalkidis, Abhik Jana, Dirk Hartung, Michael Bommarito, Ion
  Androutsopoulos, Daniel Katz, and Nikolaos Aletras.
\newblock {LexGLUE}: A benchmark dataset for legal language understanding in
  {E}nglish.
\newblock In \emph{Proceedings of the 60th Annual Meeting of the Association
  for Computational Linguistics (ACL, Volume 1: Long Papers)}, pages
  4310--4330, Dublin, Ireland, 2022. Association for Computational Linguistics.
\newblock \doi{10.18653/v1/2022.acl-long.297}.
\newblock URL \url{https://aclanthology.org/2022.acl-long.297/}.

\bibitem[Chen et~al.(2024)Chen, Xiao, Zhang, Luo, Lian, and Liu]{chen2024bgem3}
Jianlyu Chen, Shitao Xiao, Peitian Zhang, Kun Luo, Defu Lian, and Zheng Liu.
\newblock {M3-Embedding}: Multi-linguality, multi-functionality,
  multi-granularity text embeddings through self-knowledge distillation.
\newblock In \emph{Findings of the Association for Computational Linguistics:
  ACL 2024}, pages 2318--2335, 2024.
\newblock \doi{10.18653/v1/2024.findings-acl.137}.

\bibitem[Dagan et~al.(2006)Dagan, Glickman, and Magnini]{dagan2006pascal}
Ido Dagan, Oren Glickman, and Bernardo Magnini.
\newblock The {PASCAL} recognising textual entailment challenge.
\newblock In \emph{Machine Learning Challenges. Evaluating Predictive
  Uncertainty, Visual Object Classification, and Recognising Textual
  Entailment}, pages 177--190. Springer, 2006.
\newblock \doi{10.1007/11736790_9}.

\bibitem[Dalglish et~al.(2020)Dalglish, Khalid, and
  McMahon]{dalglish2020document}
Sarah~L. Dalglish, Hina Khalid, and Shannon~A. McMahon.
\newblock Document analysis in health policy research: The {READ} approach.
\newblock \emph{Health Policy and Planning}, 35\penalty0 (10):\penalty0
  1424--1431, 2020.
\newblock \doi{10.1093/heapol/czaa064}.

\bibitem[{Executive Office of the President}(2021)]{eo14008}
{Executive Office of the President}.
\newblock Executive order 14008: Tackling the climate crisis at home and
  abroad, 2021.
\newblock 86 Fed.\ Reg.\ 7619, January 27, 2021.

\bibitem[{Executive Office of the President}(2025)]{eo14154}
{Executive Office of the President}.
\newblock Executive order 14154: Unleashing american energy, 2025.
\newblock 90 Fed.\ Reg.\ 8353, January 20, 2025.

\bibitem[Freeman and Maybin(2011)]{freeman2011documents}
Richard Freeman and Jo~Maybin.
\newblock Documents, practices and policy.
\newblock \emph{Evidence \& Policy}, 7\penalty0 (2):\penalty0 155--170, 2011.
\newblock \doi{10.1332/174426411X579207}.

\bibitem[Gilardi et~al.(2023)Gilardi, Alizadeh, and Kubli]{gilardi2023chatgpt}
Fabrizio Gilardi, Meysam Alizadeh, and Ma\"{e}l Kubli.
\newblock {ChatGPT} outperforms crowd workers for text-annotation tasks.
\newblock \emph{Proceedings of the National Academy of Sciences}, 120\penalty0
  (30):\penalty0 e2305016120, 2023.
\newblock \doi{10.1073/pnas.2305016120}.
\newblock URL \url{https://doi.org/10.1073/pnas.2305016120}.

\bibitem[Goebel et~al.(2024)Goebel, Kano, Kim, Rabelo, Satoh, and
  Yoshioka]{goebel2024overview}
Randy Goebel, Yoshinobu Kano, Mi-Young Kim, Juliano Rabelo, Ken Satoh, and
  Masaharu Yoshioka.
\newblock Overview and discussion of the competition on legal information,
  extraction/entailment ({COLIEE}) 2023.
\newblock \emph{The Review of Socionetwork Strategies}, 18\penalty0
  (1):\penalty0 27--47, 2024.
\newblock \doi{10.1007/s12626-023-00152-0}.
\newblock URL \url{https://doi.org/10.1007/s12626-023-00152-0}.

\bibitem[Goebel et~al.(2025)Goebel, Kano, Kim, Kwan, Satoh, Yamada, and
  Yoshioka]{goebel2025competition}
Randy Goebel, Yoshinobu Kano, Mi-Young Kim, Calum Kwan, Ken Satoh, Hiroaki
  Yamada, and Masaharu Yoshioka.
\newblock An overview of the {COLIEE} 2025 competition: Legal case law and
  statute law information retrieval and entailment.
\newblock In \emph{Proceedings of the Twentieth International Conference on
  Artificial Intelligence and Law (ICAIL)}, pages 506--515. ACM, 2025.
\newblock \doi{10.1145/3769126.3785016}.
\newblock URL \url{https://doi.org/10.1145/3769126.3785016}.

\bibitem[Green-Pedersen(2004)]{greenpedersen2004dependent}
Christoffer Green-Pedersen.
\newblock The dependent variable problem within the study of welfare state
  retrenchment: Defining the problem and looking for solutions.
\newblock \emph{Journal of Comparative Policy Analysis: Research and Practice},
  6\penalty0 (1):\penalty0 3--14, 2004.
\newblock \doi{10.1080/1387698042000222763}.

\bibitem[Grimmer and Stewart(2013)]{grimmer2013text}
Justin Grimmer and Brandon~M. Stewart.
\newblock Text as data: The promise and pitfalls of automatic content analysis
  methods for political texts.
\newblock \emph{Political Analysis}, 21\penalty0 (3):\penalty0 267--297, 2013.
\newblock \doi{10.1093/pan/mps028}.
\newblock URL \url{https://doi.org/10.1093/pan/mps028}.

\bibitem[Gwet(2008)]{gwet2008computing}
Kilem~Li Gwet.
\newblock Computing inter-rater reliability and its variance in the presence of
  high agreement.
\newblock \emph{British Journal of Mathematical and Statistical Psychology},
  61\penalty0 (1):\penalty0 29--48, 2008.
\newblock \doi{10.1348/000711006X126600}.

\bibitem[Hacker(2004)]{hacker2004privatizing}
Jacob~S. Hacker.
\newblock Privatizing risk without privatizing the welfare state: The hidden
  politics of social policy retrenchment in the united states.
\newblock \emph{American Political Science Review}, 98\penalty0 (2):\penalty0
  243--260, 2004.
\newblock \doi{10.1017/S0003055404001121}.

\bibitem[Hall(1993)]{hall1993policy}
Peter~A. Hall.
\newblock Policy paradigms, social learning, and the state: The case of
  economic policymaking in britain.
\newblock \emph{Comparative Politics}, 25\penalty0 (3):\penalty0 275--296,
  1993.
\newblock \doi{10.2307/422246}.

\bibitem[Heilmann and Perry(2011)]{heilmann2011mao}
Sebastian Heilmann and Elizabeth~J. Perry, editors.
\newblock \emph{Mao's Invisible Hand: The Political Foundations of Adaptive
  Governance in China}.
\newblock Harvard University Asia Center, Cambridge, MA, 2011.

\bibitem[Hu et~al.(2022)Hu, Shen, Wallis, Allen-Zhu, Li, Wang, Wang, and
  Chen]{hu2022lora}
Edward~J. Hu, Yelong Shen, Phillip Wallis, Zeyuan Allen-Zhu, Yuanzhi Li, Shean
  Wang, Lu~Wang, and Weizhu Chen.
\newblock {LoRA}: Low-rank adaptation of large language models.
\newblock In \emph{International Conference on Learning Representations}, 2022.

\bibitem[Kim et~al.(2015)Kim, Goebel, and Satoh]{kim2015overview}
Mi-Young Kim, Randy Goebel, and Ken Satoh.
\newblock {COLIEE}-2015: Evaluation of legal question answering.
\newblock In \emph{Ninth International Workshop on Juris-Informatics
  (JURISIN)}, 2015.
\newblock URL \url{https://sites.ualberta.ca/~miyoung2/Papers/2015_COLIEE.pdf}.

\bibitem[Kingdon(1984)]{kingdon1984agendas}
John~W. Kingdon.
\newblock \emph{Agendas, Alternatives, and Public Policies}.
\newblock Little, Brown, Boston, 1984.

\bibitem[Koreeda and Manning(2021)]{koreeda2021contractnli}
Yuta Koreeda and Christopher~D. Manning.
\newblock {ContractNLI}: A dataset for document-level natural language
  inference for contracts.
\newblock In \emph{Findings of the Association for Computational Linguistics:
  EMNLP 2021}, pages 1907--1919, 2021.
\newblock \doi{10.18653/v1/2021.findings-emnlp.164}.

\bibitem[Laver et~al.(2003)Laver, Benoit, and Garry]{laver2003extracting}
Michael Laver, Kenneth Benoit, and John Garry.
\newblock Extracting policy positions from political texts using words as data.
\newblock \emph{American Political Science Review}, 97\penalty0 (2):\penalty0
  311--331, 2003.
\newblock \doi{10.1017/S0003055403000698}.

\bibitem[Lindblom(1959)]{lindblom1959muddling}
Charles~E. Lindblom.
\newblock The science of ``muddling through''.
\newblock \emph{Public Administration Review}, 19\penalty0 (2):\penalty0
  79--88, 1959.
\newblock \doi{10.2307/973677}.

\bibitem[{National Health Commission of the People's Republic of
  China}(2020)]{nhc2020announcement}
{National Health Commission of the People's Republic of China}.
\newblock Announcement of the national health commission of the people's
  republic of china, no.~1 of 2020.
\newblock Technical Report Announcement No.~1 of 2020, National Health
  Commission of the People's Republic of China, January 2020.
\newblock URL
  \url{https://www.nhc.gov.cn/jkj/c100063/202001/a558f073447946ae809de6da5a4b3fb9.shtml}.
\newblock Issued 20 January 2020. Classifies {COVID-19} pneumonia as a Class~B
  notifiable infectious disease subject to Class~A prevention and control
  measures. In Chinese.

\bibitem[Oi and Goldstein(2018)]{oi2018zouping}
Jean~C. Oi and Steven~M. Goldstein, editors.
\newblock \emph{Zouping Revisited: Adaptive Governance in a {C}hinese County}.
\newblock Stanford University Press, Stanford, CA, 2018.

\bibitem[Pierson(1994)]{pierson1994dismantling}
Paul Pierson.
\newblock \emph{Dismantling the Welfare State? Reagan, Thatcher, and the
  Politics of Retrenchment}.
\newblock Cambridge University Press, Cambridge, 1994.

\bibitem[Pierson(2000)]{pierson2000increasing}
Paul Pierson.
\newblock Increasing returns, path dependence, and the study of politics.
\newblock \emph{American Political Science Review}, 94\penalty0 (2):\penalty0
  251--267, 2000.
\newblock \doi{10.2307/2586011}.

\bibitem[Rabelo et~al.(2022)Rabelo, Goebel, Kim, Kano, Yoshioka, and
  Satoh]{rabelo2022overview}
Juliano Rabelo, Randy Goebel, Mi-Young Kim, Yoshinobu Kano, Masaharu Yoshioka,
  and Ken Satoh.
\newblock Overview and discussion of the competition on legal information
  extraction/entailment ({COLIEE}) 2021.
\newblock \emph{The Review of Socionetwork Strategies}, 16\penalty0
  (1):\penalty0 111--133, 2022.
\newblock \doi{10.1007/s12626-022-00105-z}.
\newblock URL \url{https://doi.org/10.1007/s12626-022-00105-z}.

\bibitem[Sabatier(1988)]{sabatier1988advocacy}
Paul~A. Sabatier.
\newblock An advocacy coalition framework of policy change and the role of
  policy-oriented learning therein.
\newblock \emph{Policy Sciences}, 21\penalty0 (2--3):\penalty0 129--168, 1988.
\newblock \doi{10.1007/BF00136406}.

\bibitem[Slapin and Proksch(2008)]{slapin2008scaling}
Jonathan~B. Slapin and Sven-Oliver Proksch.
\newblock A scaling model for estimating time-series party positions from
  texts.
\newblock \emph{American Journal of Political Science}, 52\penalty0
  (3):\penalty0 705--722, 2008.
\newblock \doi{10.1111/j.1540-5907.2008.00338.x}.

\bibitem[{State Council Joint Prevention and Control Mechanism for COVID-19,
  Comprehensive Group}(2022)]{jpcm2022newten}
{State Council Joint Prevention and Control Mechanism for COVID-19,
  Comprehensive Group}.
\newblock Notice on further optimizing and implementing {COVID-19} prevention
  and control measures.
\newblock Technical Report Lianfang Liankong Jizhi Zongfa [2022] No. 113, State
  Council of the People's Republic of China, December 2022.
\newblock URL
  \url{https://www.shanghai.gov.cn/sjzccs/20221207/90b8df6914e243a99a36c7f94a11d7ce.html}.
\newblock Issued 7 December 2022; known as the ``New Ten Measures'' (xin shi
  tiao). In Chinese.

\bibitem[Sun et~al.(2026)Sun, Chang, Ang, Mu, Xu, Zhang, and
  Hao]{sun2026capccg}
Bolun Sun, Charles Chang, Yuen~Yuen Ang, Ruotong Mu, Yuchen Xu, Zhengxin Zhang,
  and Pingxu Hao.
\newblock {CAPC-CG}: A large-scale, expert-directed {LLM}-annotated corpus of
  adaptive policy communication in {C}hina.
\newblock In \emph{Proceedings of the 64th Annual Meeting of the Association
  for Computational Linguistics (Volume 1: Long Papers)}, pages 944--966, San
  Diego, California, United States, 2026. Association for Computational
  Linguistics.
\newblock \doi{10.18653/v1/2026.acl-long.42}.
\newblock URL \url{https://aclanthology.org/2026.acl-long.42/}.

\bibitem[Weaver(1986)]{weaver1986politics}
R.~Kent Weaver.
\newblock The politics of blame avoidance.
\newblock \emph{Journal of Public Policy}, 6\penalty0 (4):\penalty0 371--398,
  1986.
\newblock \doi{10.1017/S0143814X00004219}.

\bibitem[Wilkerson and Casas(2017)]{wilkerson2017large}
John Wilkerson and Andreu Casas.
\newblock Large-scale computerized text analysis in political science:
  Opportunities and challenges.
\newblock \emph{Annual Review of Political Science}, 20\penalty0 (1):\penalty0
  529--544, 2017.
\newblock \doi{10.1146/annurev-polisci-052615-025542}.
\newblock URL \url{https://doi.org/10.1146/annurev-polisci-052615-025542}.

\bibitem[Williams et~al.(2018)Williams, Nangia, and Bowman]{williams2018broad}
Adina Williams, Nikita Nangia, and Samuel~R. Bowman.
\newblock A broad-coverage challenge corpus for sentence understanding through
  inference.
\newblock In \emph{Proceedings of the 2018 Conference of the North American
  Chapter of the Association for Computational Linguistics: Human Language
  Technologies (NAACL-HLT)}, pages 1112--1122, New Orleans, Louisiana, 2018.
  Association for Computational Linguistics.
\newblock \doi{10.18653/v1/N18-1101}.
\newblock URL \url{https://aclanthology.org/N18-1101/}.

\bibitem[Xiao et~al.(2024)Xiao, Liu, Zhang, Muennighoff, Lian, and
  Nie]{xiao2024cpack}
Shitao Xiao, Zheng Liu, Peitian Zhang, Niklas Muennighoff, Defu Lian, and
  Jian-Yun Nie.
\newblock C-pack: Packed resources for general {C}hinese embeddings.
\newblock In \emph{Proceedings of the 47th International ACM SIGIR Conference
  on Research and Development in Information Retrieval}, pages 641--649, 2024.

\bibitem[Yang et~al.(2025)]{yang2025qwen3}
An~Yang et~al.
\newblock {Qwen3} technical report.
\newblock \emph{arXiv preprint arXiv:2505.09388}, 2025.

\bibitem[Yu et~al.(2026)Yu, Li, and Ostarek]{yu2026regtrack}
Joe Yu, Kevin~Chenhao Li, and Julian Ostarek.
\newblock {R}eg{T}rack: A fine-grained benchmark for multi-class legal change
  detection.
\newblock In \emph{Proceedings of the 64th Annual Meeting of the Association
  for Computational Linguistics (Volume 4: Student Research Workshop)}, pages
  764--778, San Diego, California, United States, July 2026. Association for
  Computational Linguistics.
\newblock \doi{10.18653/v1/2026.acl-srw.68}.
\newblock URL \url{https://aclanthology.org/2026.acl-srw.68/}.

\bibitem[Zhong et~al.(2020)Zhong, Xiao, Tu, Zhang, Liu, and Sun]{zhong2020does}
Haoxi Zhong, Chaojun Xiao, Cunchao Tu, Tianyang Zhang, Zhiyuan Liu, and Maosong
  Sun.
\newblock How does {NLP} benefit legal system: A summary of legal artificial
  intelligence.
\newblock In \emph{Proceedings of the 58th Annual Meeting of the Association
  for Computational Linguistics (ACL)}, pages 5218--5230. Association for
  Computational Linguistics, 2020.
\newblock \doi{10.18653/v1/2020.acl-main.466}.
\newblock URL \url{https://aclanthology.org/2020.acl-main.466/}.

\bibitem[Ziems et~al.(2024)Ziems, Held, Shaikh, Chen, Zhang, and
  Yang]{ziems2024large}
Caleb Ziems, William Held, Omar Shaikh, Jiaao Chen, Zhehao Zhang, and Diyi
  Yang.
\newblock Can large language models transform computational social science?
\newblock \emph{Computational Linguistics}, 50\penalty0 (1):\penalty0 237--291,
  2024.
\newblock \doi{10.1162/coli_a_00502}.
\newblock URL \url{https://aclanthology.org/2024.cl-1.8/}.

\end{thebibliography}

\clearpage
\appendix

\noindent These appendices collect three kinds of material that round out the
main text: the \emph{conceptual definition} of policy change and how it can be
identified in government documents; the \emph{protocols of expert annotation},
from pair construction and the codebook to adjudication and calibration; and
the \emph{technical variations we have experimented with in eliciting,
training, and evaluating LLMs}. The appendices are ordered so that readers can
move from the conceptual and least technical material to the most technical,
with reference material at the end.

Concretely, the appendices provide (i) a conceptual account of how policy
change is communicated and why it is difficult to detect, (ii) the construction
of the training sample, (iii) the five-class codebook, (iv) a full description
of the adjudication/calibration packet, (v) the CAPC database and its ongoing
expansion, including validation on newly collected documents, (vi) the
AI-policy-change application results, (vii) a post-hoc decomposition of
five-category predictions into domain-match and direction judgments, (viii) a
comparison of joint, cascade, and direct elicitation of the dual-axis outputs,
(ix) training-regime, sample-size, rare-class, and model-family analyses,
(x) a comparison of training and test-time techniques beyond the main-paper
recipe, (xi) context-length robustness details, (xii) an evaluation of the
corpus's policy-signal (color) coding as auxiliary evidence for the direction
axis, (xiii) complete prompt listings, and (xiv) the data statement.

\section{How to Identify Policy Change in Government Documents}\label{app:identifying-change}
Policy change is seldom announced as ``policy change.'' A government may
preserve the formal shell of an earlier document while changing the policy's
scope, obligations, or implementation intensity; it may also move the same
initiative across administrative domains as priorities change. Identifying
change therefore requires more than checking whether two documents share a
topic or whether one document explicitly cites another.

\paragraph{Policy domains move.} A policy initiative does not remain inside a
fixed substantive domain. Administrative restructuring, changing political
priorities, and crisis management can move the same program across policy
areas. Zero-COVID, for example, belongs to livelihood and public services when
a document regulates lockdowns and testing, but to economic development when a
later document regulates targeted relief and recovery. A Belt-and-Road
framework can similarly move across infrastructure, finance, and security as
its instruments change. Topic labels are therefore analytical conveniences
imposed on a moving target, not stable identifiers of the policy domain.

This movement creates two failure modes for automated matching. First,
superficial topical similarity is not sufficient: two municipalities can use
near-identical administrative templates while regulating different local
matters. Second, topical dissimilarity is not necessary evidence of
irrelevance: a genuine successor can be reworded, reassigned, or reissued under
a different administrative frame. Cross-jurisdiction reissues compound the
problem because local documents often adapt a central directive without citing
the original document number (\emph{wenhao}). The first identification problem
is thus whether two texts govern the same policy domain, even when their
surface topics differ.

\paragraph{Policy change is often subtle or deliberately understated.}
Even when the policy domain is clear, the direction of change may not be.
Policymakers use flexible language to preserve local discretion, accommodate
heterogeneous conditions, and avoid publicly advertising an abrupt shift. A
restriction may be phrased as an optimization or encouragement; an expansion
may appear only as a clause that ``clarifies'' an earlier obligation; and a
reversal may be buried in a routine list of documents being cleaned up. A
single word can change the force of an obligation. Explicit repeal markers such
as \textit{f\`eizh\v{\i}} (`repeal'), \textit{sh\=ixi\`ao} (`lapse'),
\textit{zh\=ongzh\v{\i} sh\=ix\'ing} (`cease enforcement'),
\textit{b\'uz\`ai zh\'ix\'ing} (`no longer enforce'), and
\textit{t\'ingzh\v{\i} zh\'ix\'ing} (`halt enforcement') are therefore
important cues, but they are neither universal nor always displayed near the
document title.

These features make sudden policy change hard for the public to detect. The
observable signal is often a quiet change in who is covered, what is required,
which agency is responsible, or whether an earlier document remains in force.
A document-level classifier that reads only titles or topical embeddings will
miss exactly these consequential shifts. Our formulation consequently asks for
an ordered judgment between two policy directives: whether the later directive
continues, broadens, narrows, reverses, or does not govern the same domain as
the earlier one.

\paragraph{Why rare changes carry the signal.} In a natural sample of directive
pairs, continuation and irrelevance dominate; expansion, restriction, and
reversal are rare. In the first 300-pair workbook described below, 155 pairs
were consistency and none were contradiction. Yet the rare classes carry much
of the substantively important signal: they mark where obligations expand,
where coverage is withdrawn, and where a prior policy is quietly reversed. The
benchmark therefore deliberately oversamples rare change candidates, while the
main evaluation still reports performance on the full expert-labeled sample
rather than only on the enriched classes.

\section{Training-Sample Construction}\label{app:construction}
The benchmark is built from ordered pairs of \emph{policy directives}, not
from whole documents treated as indivisible units. Source documents are
segmented into policy paragraphs, each carrying a distinctive directive;
candidate pairs are then constructed from same-agency temporal sequences,
republication chains, and embedding nearest neighbors. Each model input
contains the earlier and later title, issuing agency, year, and up to 1,500
characters of directive (paragraph) text.

\paragraph{Round 1: stratified random workbook.} The first 300-pair annotation
workbook was drawn by stratified sampling across agency-match status,
temporal-gap cells, and embedding-similarity bands. This
round measured the natural difficulty of the pair frame. Its realized labels
were dominated by continuation and irrelevance: 155 consistency (51.7\%), 104
unrelated (34.7\%), 35 expansion (11.7\%), six restriction (2.0\%), and zero
contradiction.

\paragraph{Round 2: rare-category oversample.} The second 300-pair workbook
targeted the change classes that Round 1 missed. Expansion and restriction
candidates came from high- or moderate-similarity temporal pairs not already
used in Round 1. Contradiction candidates came from a corpus-wide scan for
repeal or cessation markers combined with a reference to another document's
wenhao. The scan produced 78,247 candidate contradiction pairs, from which a
stratified sample was selected for annotation.

Expert adjudication revised a substantial share of these targeted candidates.
The realized Round 2 labels were 106 unrelated (35.3\%), 68 restriction
(22.7\%), 66 expansion (22.0\%), and 60 contradiction (20.0\%); no pair was
judged consistency because consistency candidates were not targeted. The
oversample thus supplied 194 change-class pairs, compared with 41 in Round 1,
while also demonstrating why candidate generation cannot be treated as a
labeled resource: roughly one-third of targeted candidates were unrelated on
expert inspection.

\paragraph{Targeted top-ups and final gold set.} Later rounds expanded the
scarcest categories, especially restriction and contradiction. Round 3 added
226 pairs (98 hard cases, 101 restriction candidates, and 27 contradiction
candidates), and Round 4 added 174 structurally enriched rare-class pairs. The
union of the four rounds gives the final 1,000-pair gold set: 255 consistency,
171 expansion, 176 restriction, 127 contradiction, and 271 unrelated.

\paragraph{Training and evaluation splits.} The modeling sample preserves the
ordered pair as the unit of analysis. Cross-validation uses five folds that are
both label-stratified and source-document-group disjoint: if two pairs share a
source document, they remain in the same fold. The canonical 80/20 run reported
in Appendix~\ref{app:techniques} uses a separate seed and is reported apart
from the fold-0 technique comparison.

\paragraph{Existing-label and frame-sensitivity analyses.} We first use the
initial 300-pair, less-enriched workbook to test whether the main result is
specific to the later rare-class top-ups. Table~\ref{tab:early-frame} shows
that expert fine-tuning improves over zero-shot models on this earlier frame
as well. The frame is stratified rather than uniformly natural, so we report
it as a robustness subset, not as a natural-distribution estimate.

\begin{table}[t]
\centering
\small
\caption{Existing-model results on the initial 300-pair, less-enriched frame. m-F1 obs.\ averages over classes present in the frame's gold labels; binary $\kappa$ measures related versus irrelevant. This stratified frame is not a uniform natural sample. SFT predictions are out-of-fold.}
\label{tab:early-frame}
\begin{tabular}{lrrrrr}
\toprule
System & $n$ & m-F1 & m-F1 obs. & $\kappa$ & binary $\kappa$ \\
\midrule
Qwen3-32B zero-shot & 300 & 0.403 & 0.503 & 0.411 & 0.357 \\
GPT-5 zero-shot & 300 & 0.329 & 0.411 & 0.240 & 0.307 \\
Five-way SFT & 300 & 0.474 & 0.593 & 0.581 & 0.665 \\
Five-way SFT jurcorr & 300 & 0.425 & 0.532 & 0.479 & 0.520 \\
Dual-axis SFT & 300 & 0.542 & 0.677 & 0.638 & 0.702 \\
Dual-axis SFT jurcorr & 300 & 0.481 & 0.602 & 0.484 & 0.539 \\
\bottomrule
\end{tabular}
\end{table}

Two further analyses require no new annotation. For the embedder question,
Table~\ref{tab:embedding-robustness} compares BGE-large-zh and
Qwen3-Embedding-8B on the already annotated gold pairs, including cosine
bands and true-partner retrieval ranks. Existing-model scores by band remain
human-referenced because the embedder changes only the diagnostic strata,
not the labels. Retrieval is bounded to the 1,961 benchmark source documents
and is not a full-corpus recall estimate.

\begin{table}[t]
\centering
\small
\caption{Alternate-embedder diagnostics on the existing 1,000 gold pairs. Retrieval ranks are computed within the 1,961 benchmark source documents, not the full corpus. On 520 comparable non-sentinel PKUlaw pairs, original and recomputed BGE cosines have Spearman 1.000 and median absolute difference 0.0000.}
\label{tab:embedding-robustness}
\begin{tabular}{lrr}
\toprule
Metric & BGE-large-zh & Qwen3-Embedding-8B \\
\midrule
Gold-partner recall, top 10 & 87.3\% & 89.9\% \\
Gold-partner recall, top 50 & 93.7\% & 95.2\% \\
Median gold-partner rank & 2.0 & 2.0 \\
\midrule
Cosine Spearman, BGE versus Qwen & \multicolumn{2}{c}{0.823} \\
Similarity-band agreement & \multicolumn{2}{c}{40.4\%} \\
\bottomrule
\end{tabular}
\end{table}

The two embedders rank cosines similarly overall and attain comparable top-50
partner recall, but agree on the exact similarity band for only 40.4\% of
pairs. Table~\ref{tab:embedding-model-bands} therefore reports existing-model
scores under both bandings. Fine-tuning gains persist in the high, moderate,
and low bands under both embedders; the Qwen below-band is the lone macro-F1
exception, illustrating that absolute scores remain stratum-dependent.

\begin{table}[t]
\centering
\small
\caption{Macro-F1 by embedding-similarity band on the same human-labeled gold pairs. Bands are diagnostics, not new train/test splits. The BGE below-band contains one pair and is not interpretable as a performance estimate.}
\label{tab:embedding-model-bands}
\begin{tabular}{llrrrr}
\toprule
Embedder & Band & $n$ & Qwen3-32B & Five-way SFT & Dual-axis SFT \\
\midrule
BGE & high & 443 & 0.465 & 0.629 & 0.639 \\
BGE & moderate & 378 & 0.358 & 0.553 & 0.570 \\
BGE & low & 178 & 0.420 & 0.603 & 0.677 \\
BGE & below & 1 & 0.200 & 0.200 & 0.200 \\
\midrule
Qwen & high & 144 & 0.441 & 0.585 & 0.577 \\
Qwen & moderate & 407 & 0.449 & 0.586 & 0.591 \\
Qwen & low & 380 & 0.390 & 0.581 & 0.626 \\
Qwen & below & 69 & 0.412 & 0.405 & 0.365 \\
\bottomrule
\end{tabular}
\end{table}

Table~\ref{tab:frame-composition} separately compares a uniform natural-pair
frame (1,000 pairs from a fixed 20,000-document universe) with matched
1,200-pair BGE and Qwen retrieval frames. Only 16 pairs occur in both matched
frames, and their predicted class compositions differ. The classifier's
outputs describe predicted frame composition only; they are not treated as
human-validated accuracy. These are deliberately bounded diagnostics of
sampling sensitivity, distinct from the human-labeled early-frame and
strict-recode results.

\begin{table}[t]
\centering
\small
\caption{Predicted composition of the natural and matched retrieval frames. Directional pairs are expansion, restriction, or reversal. These are model predictions for sampling-sensitivity analysis, not human-validated accuracy. The BGE and Qwen frames overlap on 16 pairs; their cosines have Spearman correlation 0.873 across the alternate-frame union.}
\label{tab:frame-composition}
\begin{tabular}{lrrrrr}
\toprule
Frame & $n$ & Irrelevant & Consistency & Directional & Mean conf. \\
\midrule
Uniform natural & 1000 & 99.8\% & 0.2\% & 0.0\% & 1.000 \\
BGE retrieval & 1200 & 66.4\% & 30.6\% & 3.0\% & 0.996 \\
Qwen retrieval & 1200 & 47.4\% & 50.4\% & 2.2\% & 0.996 \\
\bottomrule
\end{tabular}
\end{table}

\paragraph{Why agency/year constraints are necessary.} Unrestricted candidate
generation would ask whether directives issued by unrelated agencies, across
provinces and distant years, govern the same domain; most such pairs do not.
Agency and temporal constraints are therefore substantive domain-matching
controls, not arbitrary filters. Table~\ref{tab:agency-year-design} supports
this design with two kinds of evidence. In the early human-labeled frame,
18.5\% of different-agency pairs are related, against 70.1\% of same-agency
pairs; directional shares are 0.0\% and 14.8\%, respectively. The frame is
stratified, so these are yields within design cells, not natural prevalence.
The unrestricted uniform frame has only 0.2\% predicted-related pairs, against
41.7\% and 57.3\% in the same-agency/1--5-year BGE and Qwen frames. Those
latter figures are composition diagnostics, not model-generated truth. These
constraints guide candidate retrieval only; they never determine the relation
label, and cross-agency or cross-year pairs can still be related.

\begin{table}[t]
\centering
\small
\caption{Agency/year constraints increase the yield of related candidate pairs. The top panel uses existing human labels from the stratified early frame and is not a corpus-prevalence estimate; nine agency-missing rows are excluded from the same/different-agency comparison. The bottom panel reports model-predicted composition only, not human-validated accuracy.}
\label{tab:agency-year-design}
\begin{tabular}{llrrr}
\toprule
Evidence & Pair selection & $n$ & Related & Directional \\
\midrule
Human labels & Different agencies & 27 & 18.5\% & 0.0\% \\
Human labels & Same agency & 264 & 70.1\% & 14.8\% \\
Human labels & Other agency/year cells & 72 & 48.6\% & 9.7\% \\
Human labels & Same agency, 1--5 years & 228 & 70.6\% & 14.9\% \\
\midrule
Model composition & Uniform natural, unrestricted & 1000 & 0.2\% & 0.0\% \\
Model composition & BGE retrieval, same agency, 1--5 years & 600 & 41.7\% & 1.7\% \\
Model composition & QWEN retrieval, same agency, 1--5 years & 600 & 57.3\% & 2.0\% \\
\bottomrule
\end{tabular}
\end{table}

\section{Five-Class Relation Codebook}\label{app:codebook}
Table~\ref{tab:codebook} reproduces the relation definitions given to expert
annotators. The shared-domain gate is intentionally permissive: a shared
program, regulatory regime, or policy initiative can support a relation even
when the documents use different terminology, while shared formulaic language
without a shared substantive domain does not.

\begin{table}[!htbp]\centering\footnotesize
\setlength{\tabcolsep}{5pt}
\begin{tabular}{l p{11.2cm}}
\toprule
\textbf{Label} & \textbf{Definition and worked example} \\
\midrule
CONSISTENCY & Policy $B$ carries forward $A$ with no material change in scope
or stringency, including renewals, transmissions, and parallel reports.
\newline \textit{Example.} $A$ (2012): ``Xuzhou Human Resources and Social
Security Bureau, notice publishing the 2012 work-injury insurance data and the
periodic benefit-adjustment measures''; $B$ (2016): the same notice issued for
2016. The instrument, coverage, and obligations are unchanged; only the
reference year moves. \\
\addlinespace
EXPANSION & $B$ broadens $A$: more coverage, more obligations, or more
stringent requirements. \newline \textit{Example.} $A$ (2003): ``Ministry of
Water Resources, opinions on strengthening the management of village and
township water-supply projects''; $B$ (2004): ``Ministry of Water Resources,
notice issuing the \emph{Staffing Standards for Village and Township
Water-Supply Stations}''. $B$ adds concrete staffing obligations to the
program $A$ established. \\
\addlinespace
RESTRICTION & $B$ narrows $A$: removes coverage, relaxes obligations, or
reduces stringency. \newline \textit{Example.} $A$ (2007): ``Guangdong Price
Bureau and provincial development-reform and economic-trade commissions,
notice on further implementing the differential electricity-pricing policy for
the steel industry''; $B$ (2011): the same agencies' ``reply approving the
cessation of differential electricity pricing for Wenrong Steel Mill in Zijin
County, Heyuan City, and other enterprises''. The policy stands, but named
enterprises are removed from its scope. \\
\addlinespace
CONTRADICTION & $B$ rescinds or reverses material provisions of $A$, often via
an explicit \textit{f\`eizh\v{\i}} (`repeal') marker citing $A$'s wenhao or
policy-document ID. \newline \textit{Example.} $A$ (2020): ``L\"uliang
Municipal People's Government, opinions on accelerating enterprise listing and
share-quotation work''; $B$ (2025): ``L\"uliang Municipal People's Government,
notice repealing certain documents'', which lists $A$ among the repealed
items. \\
\addlinespace
UNRELATED & $A$ and $B$ are not substantively related; topical-similarity-only
matches fall here. \newline \textit{Example.} $A$ (1993): ``Jilin City interim
measures for the administration of urban piped gas''; $B$ (2013): ``Benxi City
measures for gas administration''. The vocabulary and policy area are similar,
but two municipalities regulate their own gas systems, so there is no shared
policy domain. \\
\bottomrule
\end{tabular}
\caption{The five-class relation label set. Examples are actual benchmark
pairs, with titles translated from Chinese. ``Reversal'' in the main text is
the prose description of the stored \textsc{contradiction} label.}
\label{tab:codebook}
\end{table}

A single source document can contain multiple clauses and actions. To avoid
mixing unrelated actions inside one annotation unit, the benchmark operates on
policy directives (policy paragraphs) rather than whole documents. Annotators
were instructed to code shifts conservatively: expansion, restriction, or
contradiction is assigned only when the two directives clearly differ in scope,
coverage, or implementation intensity. Name changes and unclear discrepancies
do not by themselves constitute a shift. This conservative rule can omit some
attempted local changes, but it makes the recorded changes more credible.

\section{The Adjudication Packet}\label{app:adjudication}
The calibration exercise described in the main paper consists of four
artifacts, distributed to two annotators who had not seen the finalized
codebook:
\begin{enumerate}
\item a \textbf{training guide} stating the shared-policy-domain gate, the
reversal cue list, the graded-class distinctions with worked examples, and
the mixed-change (single-label ladder / conservative-shift) conventions,
together with the disagreement statistics from the first reliability round;
\item a \textbf{90-pair training set} containing both coders' original
labels, the gold label, and a written rationale per pair;
\item a \textbf{30-pair blind calibration test} (texts only) on which each
annotator is scored against the other annotator and against held-out gold,
both before and after training; and
\item a \textbf{pair-detail table} with cue positions used internally for
error analysis.
\end{enumerate}
Agreement is re-measured after calibration on the blind test, so the effect
of adjudication is itself a measured quantity rather than an assumption.
Table~\ref{tab:blind-test} reports the returned post-calibration tests.
The two newly trained coders converge on each other at five-way
$\kappa = 0.519$ (binary relatedness $\kappa = 0.667$)---the same level as
the original coder-mutual ceiling, on a deliberately harder, boundary-loaded
test. Their agreement with the first-pass gold is low, and informatively so:
the test contains no gold-irrelevant pairs by construction, and 16 of its 30
pairs are precisely the contested boundary cases on which the strict
codebook and the liberal first-pass protocol diverge. Calibration therefore
produces inter-coder convergence on the strict gate; it does not by itself
adjudicate the gate, which is the subject of the next subsection.

\begin{table}[h]
\centering
\footnotesize
\caption{Post-calibration blind test ($n=30$, boundary-loaded by design).}
\label{tab:blind-test}
\begin{tabular}{lrrrr}
\toprule
Comparison & \multicolumn{2}{c}{Five-way} & \multicolumn{2}{c}{Binary relatedness} \\
\cmidrule(lr){2-3}\cmidrule(lr){4-5}
 & acc. & $\kappa$ & acc. & $\kappa$ \\
\midrule
Coder C vs gold & 0.167 & 0.038 & 0.500 & 0.000 \\
Coder D vs gold & 0.133 & 0.008 & 0.533 & 0.000 \\
Coder C vs Coder D (mutual) & 0.667 & 0.519 & 0.833 & 0.667 \\
\bottomrule
\end{tabular}

\end{table}

\subsection{Strict-Rule Recode and Boundary Analysis}
After the calibration exercise, 173 unique pairs were recoded under the strict
rule with a written rationale for every nonblank decision. Of these, 163 belong
to the 1,000-pair benchmark and ten are additional central-document pairs.
The released instrument preserves earlier and strict labels, all rationales,
and full directive texts, without exposing model predictions. A 26-pair
priority list identifies five missing labels, 17 valid cross-round revisions,
and four round-5 conflicts (one revised pair also conflicts across recodes).
We leave these 26 flagged rather than assigning them unvalidated labels.

The 117 nonblank labels in the 120-pair recode provide a strict-rule human
reference for a no-new-annotation sensitivity analysis. Agreement with this
reference is low across systems, confirming the difficulty of the boundary.
The original gold agrees with the strict reference at accuracy 0.265 and
$\kappa = 0.082$; jurisdiction-corrected five-way SFT reaches 0.393 and 0.168,
while GPT-5 reaches 0.453 and 0.255. Thus some apparent model errors are
protocol-boundary disagreements, and the strict protocol can also change model
rankings. We report rather than suppress that sensitivity, but do not treat a
single-coder recode as independently adjudicated gold or as a natural sample.

\begin{table}[t]
\centering
\small
\caption{Existing-model agreement with the 117 nonblank strict-rule labels in the 120-pair recode. This single-coder, boundary-loaded reference is not independently adjudicated gold and not a natural sample. SFT predictions are out-of-fold.}
\label{tab:strict120}
\begin{tabular}{lrrrrr}
\toprule
System & $n$ & m-F1 & m-F1 obs. & $\kappa$ & binary $\kappa$ \\
\midrule
Qwen3-32B zero-shot & 117 & 0.366 & 0.366 & 0.109 & 0.056 \\
GPT-5 zero-shot & 117 & 0.367 & 0.367 & 0.255 & 0.429 \\
Five-way SFT & 117 & 0.274 & 0.274 & 0.113 & 0.295 \\
Five-way SFT jurcorr & 117 & 0.296 & 0.296 & 0.168 & 0.362 \\
Dual-axis SFT & 117 & 0.251 & 0.251 & 0.074 & 0.213 \\
Dual-axis SFT jurcorr & 117 & 0.282 & 0.282 & 0.142 & 0.339 \\
\bottomrule
\end{tabular}
\end{table}

\subsection{Jurisdiction Correction of the Gold Set}
A jurisdiction audit of the full gold set identified 94 cross-jurisdiction
pairs and 64 pairs whose jurisdiction relation could not be parsed
automatically. Manual adjudication of all 158 confirmed 56 as legitimately
related and flipped 60 gold labels to irrelevant: a later document from one
prefecture cannot revise another prefecture's policy, whereas a province and
its own prefectures form an eligible hierarchy. The three flipped pairs
inside the blind test are exactly the pairs both coders had independently
called irrelevant. Table~\ref{tab:jurcorr-rescore} re-scores stored systems
against the corrected gold without any new inference; the main paper's
fine-tuning section then reports the classifiers retrained on the corrected
labels, which is the like-for-like comparison.

\begin{table}[h]
\centering
\footnotesize
\setlength{\tabcolsep}{4pt}
\caption{Systems re-scored on original versus jurisdiction-corrected gold
($n=1{,}000$), without retraining.}
\label{tab:jurcorr-rescore}
\begin{tabular}{lrrrr}
\toprule
System & \multicolumn{2}{c}{Original gold} & \multicolumn{2}{c}{Corrected gold} \\
\cmidrule(lr){2-3}\cmidrule(lr){4-5}
 & macro-F1 & $\kappa$ & macro-F1 & $\kappa$ \\
\midrule
Qwen3-32B (zero-shot) & 0.452 & 0.337 & 0.433 & 0.305 \\
GPT-5 (zero-shot) & 0.421 & 0.256 & 0.423 & 0.261 \\
Gemini 2.5 Pro (zero-shot) & 0.419 & 0.274 & 0.435 & 0.303 \\
Claude Sonnet 4.5 (zero-shot) & 0.412 & 0.265 & 0.416 & 0.271 \\
DeepSeek-V3 (zero-shot) & 0.406 & 0.282 & 0.412 & 0.293 \\
SFT five-way (trained on original gold) & 0.634 & 0.524 & 0.626 & 0.505 \\
SFT dual-axis (trained on original gold) & 0.649 & 0.548 & 0.635 & 0.524 \\
\bottomrule
\end{tabular}

\end{table}

\begin{table*}[t]
\centering
\footnotesize
\setlength{\tabcolsep}{4pt}
\caption{Five-fold results on the jurisdiction-corrected gold set
($n=1{,}000$). Original-gold-trained models are re-scored without
retraining; retrained models use identical folds and hyperparameters.}
\label{tab:jurcorr-retrained}
\resizebox{\textwidth}{!}{%
\begin{tabular}{lrrrrrrrrr}
\toprule
Model & $n$ & macro-F1 & $\kappa$ & acc. & F1 irrelevant & F1 consistency & F1 restriction & F1 expansion & F1 reversal \\
\midrule
Five-way SFT, original-gold trained & 1000 & 0.626 & 0.505 & 0.613 & 0.581 & 0.670 & 0.534 & 0.552 & 0.793 \\
Five-way SFT, retrained on corrected gold & 1000 & 0.634 & 0.529 & 0.637 & 0.675 & 0.690 & 0.488 & 0.527 & 0.790 \\
Dual-axis SFT, original-gold trained & 1000 & 0.635 & 0.524 & 0.624 & 0.590 & 0.685 & 0.565 & 0.546 & 0.789 \\
Dual-axis SFT, retrained on corrected gold & 1000 & 0.643 & 0.540 & 0.642 & 0.684 & 0.659 & 0.531 & 0.557 & 0.782 \\
\bottomrule
\end{tabular}
}
\end{table*}

\paragraph{Multi-label changes.} Some documents change scope and stringency
in opposite directions simultaneously. The codebook resolves these through a
single-label ladder with a conservative-shift rule; one improvised composite
label observed during annotation was adjudicated as a miscoding and is not in
the gold set. A multi-label variant of the task is future work.

\section{The CAPC Database and Its Ongoing Expansion}\label{app:capc-versions}
This appendix explains how the published CAPC corpus relates to the much
larger working corpus, describing each corpus frame by what it contains
rather than by internal version labels.
Table~\ref{tab:capc_versions} separates four frames: the \emph{published
corpus} (CAPC-CG, released with Sun et al.\ 2026), which covers central party
and government directives from 1949--2023; our \emph{extended central
snapshot}, which adds ministerial document types through 2024 and serves as
the pair-mining frame for the benchmark; an \emph{issuer-matched universe}
assembled for a separate issuer-coverage study; and the \emph{integrated
multi-level corpus}, the ongoing collection effort that combines a leading
commercial policy database with a government-website scrape across national,
ministerial, provincial, prefectural, and county levels. After verified exact-title
overlap removal, the integrated corpus core contains 4,997,123 documents.

\begin{table}[!htbp]\centering\small
\caption{The CAPC database and its ongoing expansion. Paragraph counts use each frame's own segmentation and are not cross-comparable; compare documents.}\label{tab:capc_versions}
\resizebox{\textwidth}{!}{%
\begin{tabular}{llrrll}\toprule
frame & scope & documents & paragraphs & years \\
\midrule
Published corpus (CAPC-CG) & central only & 337,038 & 3,275,477 & 1949–2023 \\
Extended central snapshot & central only & 368,679 & 4,101,064 & 1949–2024 \\
Issuer-matched universe & central + party organs & 456,673 & 3.16M (corpus seg.) & within 1921–2026 \\
Integrated multi-level corpus (ongoing) & all administrative levels & 4,997,123 & 32.46M uids + 3.54M govweb (segs. differ) & 1921–2026 \\
\bottomrule\end{tabular}}%
\end{table}

Table~\ref{tab:capc3_components} reports the component registry of the
integrated corpus.
Only the commercial database backbone and government-website corpus enter the core-text
total. The gov.cn and FLK collections are metadata registers; leader-speech
corpora are separate communication channels; Weibo and WeChat are excluded
because no complete usable corpus has been collected; and the message-board /
12345 data are demand-side petitions and replies, not government policy
documents.

\begin{table}[!htbp]\centering\small
\caption{Components of the ongoing collection effort.}\label{tab:capc3_components}
\resizebox{\textwidth}{!}{%
\begin{tabular}{llrll}\toprule
component & status & documents & years & levels \\
\midrule
Commercial policy database (backbone) & core & 4,842,217 & 1921–2026 & national–county \\
Government-website scrape & core & 159,135 & 1993–2026 & national–county (pref.-heavy) \\
gov.cn policy register (full pull) & register & 120,355 & 1996–2026 & — \\
gov.cn policy register (policy-labelled slice) & register & 39,091 & 1996–2026 & — \\
gov.cn State Council gazette register & register & 15,141 & 2000–2026 & — \\
National Laws and Regulations Database register & register & 24,280 & 2000–2025 & — \\
Leader speeches (Deng/Jiang/Hu) & channel & 620 & 1938–2012 & — \\
Xi Jinping speeches/articles & channel & 14,332 & 2012–2025 & — \\
WeChat official accounts & excluded & — & — & — \\
Weibo government accounts & excluded & 702 & — & — \\
Message-board / 12345 hotline & excluded & — & — & — \\
\bottomrule\end{tabular}}%
\end{table}

Finally, Table~\ref{tab:central_vs_all} shows the substantive difference that
matters for this paper: the extended central snapshot is central-only by
construction, whereas the
integrated corpus is overwhelmingly provincial and prefectural. This is why
the benchmark's
pair-mining frame is not interchangeable with the full working corpus, and why
we report the candidate-generation frame explicitly rather than treating the
1,000-pair benchmark as a prevalence estimate for all Chinese policy
communication.

\begin{table}[!htbp]\centering\small
\caption{The central-only extended snapshot versus the all-levels integrated corpus (documents by administrative level). The government-website scrape is shown separately because its exact-title overlap with the backbone is subtracted only in the deduplicated total.}\label{tab:central_vs_all}
\resizebox{\textwidth}{!}{%
\begin{tabular}{lrrrrrrr}\toprule
corpus & national & ministerial & provincial & prefectural & county & unknown & central (\%) \\
\midrule
Extended central snapshot & 13,696 & 353,368 & 0 & 0 & 0 & 1,615 & 100.0 \\
Commercial database backbone (labelled) & 19,010 & 1,069,449 & 2,066,418 & 1,239,252 & 88,253 & 296,592 & 22.8 \\
Government-website scrape & 40 & 1,134 & 28,453 & 126,945 & 2,563 & 0 & 0.7 \\
Integrated corpus (gross) & 19,050 & 1,070,583 & 2,094,871 & 1,366,197 & 90,816 & 296,592 & 22.1 \\
\bottomrule\end{tabular}}%
\end{table}

\begin{figure}[t]
\centering
\includegraphics[width=0.9\columnwidth]{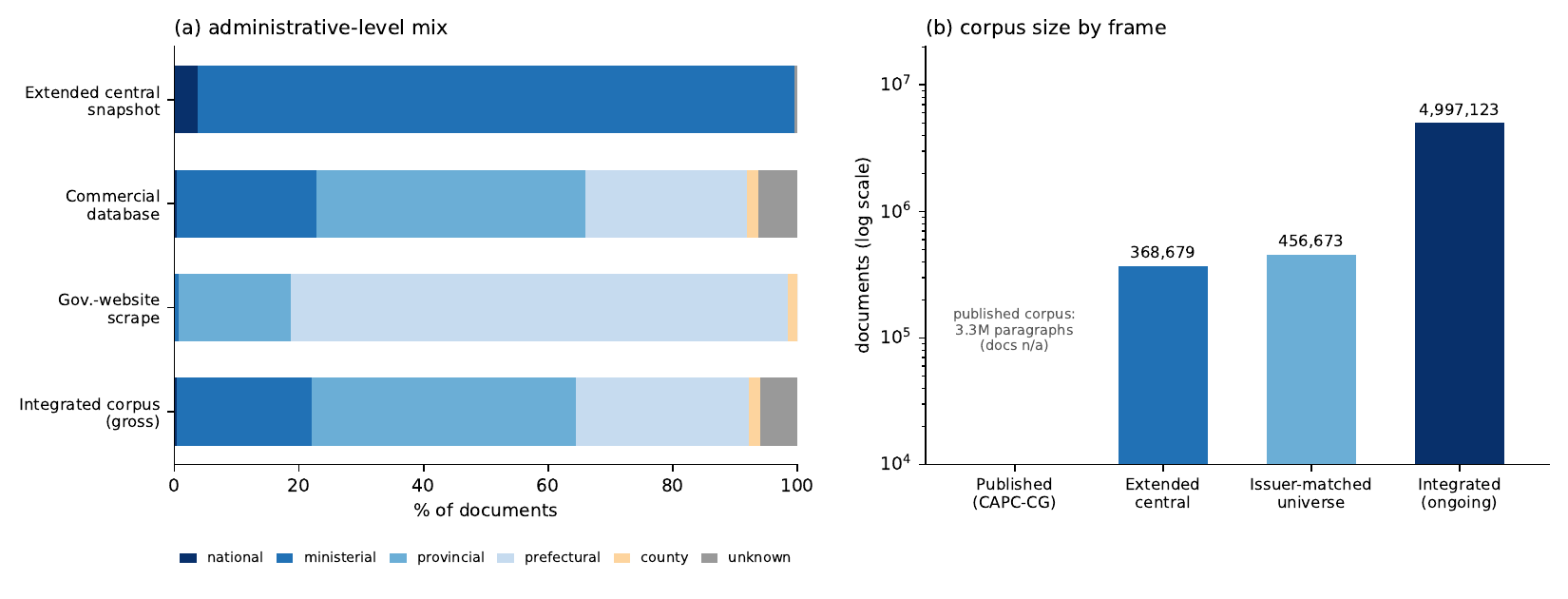}
\caption{Administrative-level composition of the extended central snapshot,
the commercial database backbone, the government-website corpus, and the integrated
multi-level corpus.}
\label{fig:capc-levels}
\end{figure}

\subsection{Validating the Classifier on Newly Collected Documents}
\label{app:govweb-validity}
The ongoing expansion is scientifically useful only if the trained classifier
keeps working on documents collected after the benchmark was built. We
therefore apply the five fine-tuned folds (majority vote) to 20,000 pairs
drawn from the government-website corpus---documents collected from live
government portals, 90\% of them dated 2024 or later and thus absent from
both the published corpus and the benchmark's pair-mining frame. Pairs come
from three channels: same-agency temporal neighbours, reissue chains (same
normalized title), and repeal-cue pairs (cessation language in the later
document). All 20,000 pairs were classified (five-of-five fold parse
coverage; mean ensemble confidence 0.931; 77.6\% unanimous).

\begin{table}[t]
\centering
\small
\caption{Predicted five-way change categories for document pairs drawn
from newly collected Chinese government-website documents (2015--2026),
by pairing channel. \emph{Directional} = expansion + restriction +
reversal. Labels are the majority vote of five fine-tuned classifier
folds applied to the first 1{,}500 characters of each document.}
\label{tab:govweb-validity}
\begin{tabular}{lrrrrrrr}
\toprule
Channel & $n$ & Unrel. & Consist. & Expans. & Restr. & Reversal & Direct. \\
\midrule
Same-agency neighbours & 14,929 & 74.4 & 21.9 & 3.4 & 0.3 & 0.0 & 3.7 \\
Reissue chains & 2,326 & 1.6 & 78.0 & 13.8 & 6.6 & 0.0 & 20.4 \\
Repeal-cue pairs & 2,745 & 64.4 & 19.0 & 7.9 & 3.5 & 5.2 & 16.6 \\
All pairs & 20,000 & 64.5 & 28.0 & 5.2 & 1.5 & 0.7 & 7.4 \\
\bottomrule
\end{tabular}
\end{table}

Table~\ref{tab:govweb-validity} shows that the classifier recovers
directional change in this new frame at the rates the channel design
predicts: reissue chains are mostly continuations (78.0\%) with a substantial
directional minority (20.4\%); repeal-cue pairs yield 16.6\% directional
change, including nearly all detected reversals; and same-agency neighbours
behave as the intended low-yield baseline (3.7\%). Repeal-cue pairs whose
automatically matched earlier document is a weak title match are mostly
unrelated (82.2\%), confirming once more that candidate generation supplies
candidates, not labels. By later-document year, the directional share is
13.4\% (2024), 10.6\% (2025), and 5.3\% (2026); the 2026 dilution reflects
the large volume of routine municipal procurement and announcement material
collected that year, while the chain and repeal channels hold up throughout
(31.5\% and 20.9\% directional in 2024).

As a human validity check, we hand-read ten high-confidence directional
detections (three reversals, three restrictions, four expansions, titles and
decisive phrases translated): eight are clearly correct, and the two
exceptions are data artifacts rather than detection failures---a
repealed document reposted by a portal with a later date, which inverts the
pair chronology, and a repeal notice at the restriction/reversal boundary.
The substantive point is that these changes are observable at all only
because collection continued: they live in prefectural and municipal
documents issued after 2023, outside both the published central corpus and
the benchmark frame. Continued multi-level collection is therefore not just
a larger corpus but the precondition for measuring current policy change.

\section{AI-Policy Change in China, 1980--2024}\label{app:aipolicy}
The published CAPC-CG corpus documents central-government policy communication
through 2023 and identifies multi-level communication as a natural extension.
Our AI-policy application uses that central corpus as the reference frame and
then asks what changes when the integrated multi-level corpus adds provincial
implementation documents. We apply the best cross-validated classifier to
AI-related policy pairs formed from same-agency temporal sequences,
republication chains, and embedding nearest neighbors among AI-flagged
documents (administrative levels 1--3, 1980--2024). We report the predicted
change-category distribution by era (pre-1993, 1993--2002, 2003--2012,
2013--2024) with per-cell counts, a timeline figure, and a spot-checked
sample of ten pairs across categories with their decisive text spans.

\begin{table*}[t]
\centering
\footnotesize
\caption{Predicted five-way change categories for AI-policy pairs by era.}
\label{tab:ai-change-era}
\resizebox{\textwidth}{!}{%
\begin{tabular}{lrrrrrr}
\toprule
Era (by later doc) & Consistency & Expansion & Restriction & Reversal & Irrelevant & $n$ \\
\midrule
Deng & -- & -- & -- & -- & -- & 0 \\
Jiang & 7.7\% (2) & 0.0\% (0) & 0.0\% (0) & 0.0\% (0) & 92.3\% (24) & 26 \\
Hu & 17.9\% (128) & 5.0\% (36) & 1.4\% (10) & 0.0\% (0) & 75.7\% (542) & 716 \\
Xi & 4.8\% (3109) & 1.8\% (1147) & 0.1\% (86) & 0.0\% (18) & 93.3\% (60951) & 65311 \\
\bottomrule
\end{tabular}%
}
\end{table*}

\begin{figure}[t]
\centering
\includegraphics[width=\columnwidth]{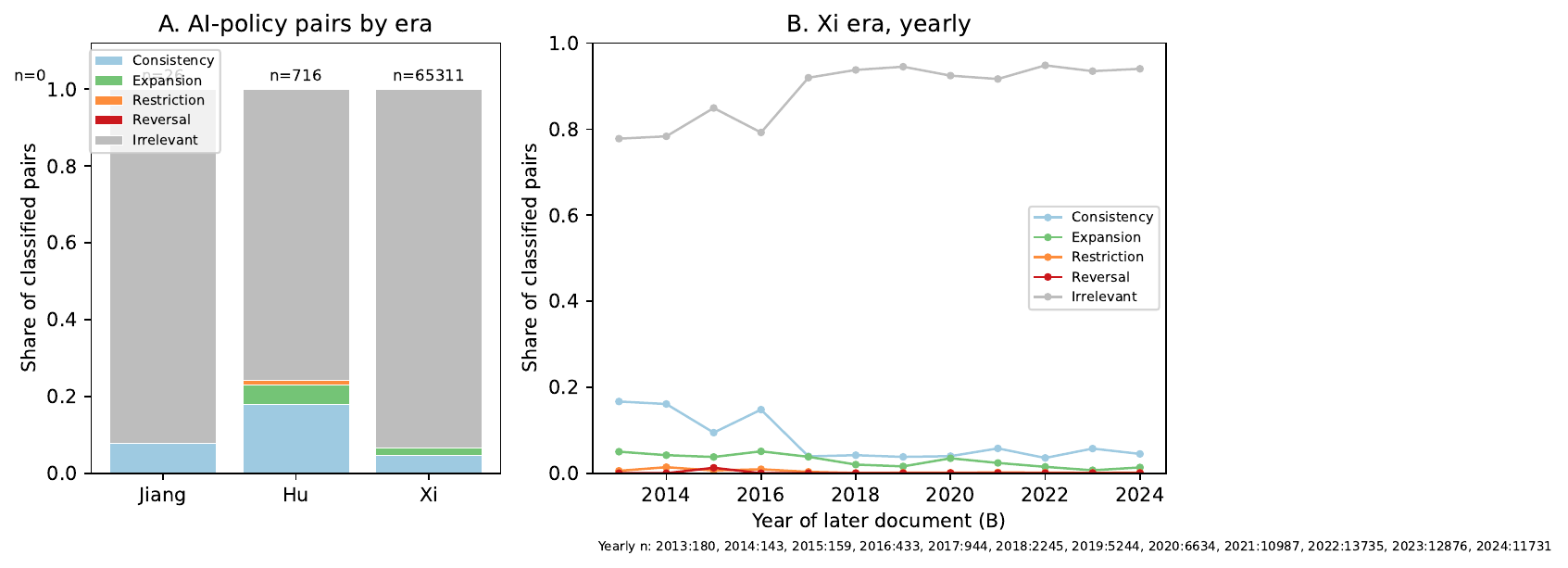}
\caption{Predicted five-way change categories for AI-policy pairs by era and,
for the Xi era, by year.}
\label{fig:ai-change-era}
\end{figure}

\begin{figure}[t]
\centering
\includegraphics[width=0.9\columnwidth]{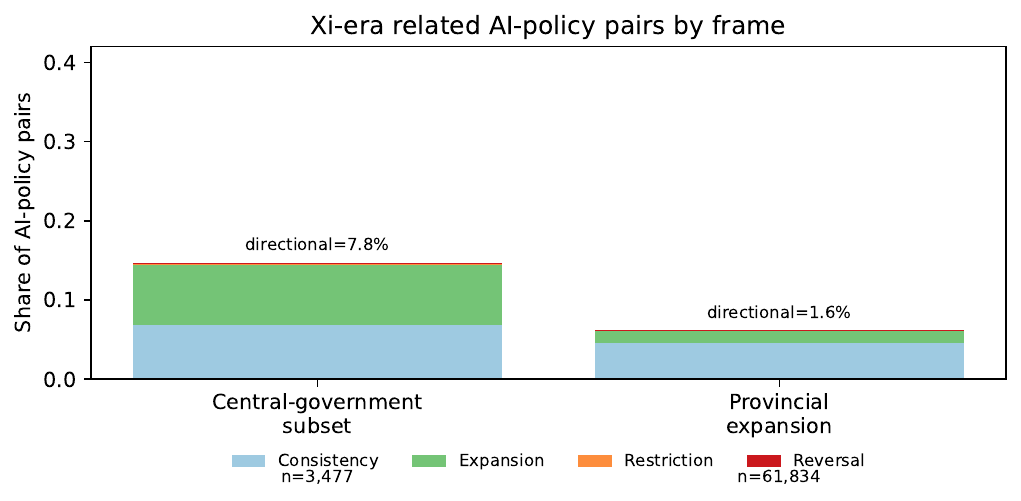}
\caption{Xi-era related AI-policy pairs in the central-government subset and
the provincial expansion of the integrated corpus.}
\label{fig:ai-central-provincial}
\end{figure}

\paragraph{Central reference frame versus provincial expansion.}
Table~\ref{tab:ai-central-provincial} splits the AI-policy frame into the
central-government subset (national and ministerial documents, comparable to
the published corpus in coverage) and the
provincial expansion made visible by the integrated corpus. Directional
AI-policy changes
are more frequent in the central frame than in the provincial frame,
especially under Xi, while the provincial expansion contributes the vast
majority of pairs. Figure~\ref{fig:ai-central-provincial} summarizes the
Xi-era related-pair distribution.

\begin{table}[!htbp]
\centering
\footnotesize
\caption{AI-policy pair classification in the central-government subset versus the provincial expansion of the integrated corpus. Shares are within frame and era; directional change is expansion + restriction + reversal.}
\label{tab:ai-central-provincial}
\resizebox{\textwidth}{!}{%
\begin{tabular}{llrrrrrr}
\toprule
Frame & era & $n$ & unrelated & consistency & expansion & restriction & reversal \\
\midrule
Central-government subset & Jiang & 1 & 100.0\% & 0.0\% & 0.0\% & 0.0\% & 0.0\% \\
Central-government subset & Hu & 197 & 67.0\% & 22.8\% & 7.6\% & 2.5\% & 0.0\% \\
Central-government subset & Xi & 3477 & 85.3\% & 6.9\% & 7.6\% & 0.1\% & 0.1\% \\
Central-government subset & All & 3675 & 84.3\% & 7.7\% & 7.6\% & 0.2\% & 0.1\% \\
Provincial expansion & Jiang & 25 & 92.0\% & 8.0\% & 0.0\% & 0.0\% & 0.0\% \\
Provincial expansion & Hu & 519 & 79.0\% & 16.0\% & 4.0\% & 1.0\% & 0.0\% \\
Provincial expansion & Xi & 61834 & 93.8\% & 4.6\% & 1.4\% & 0.1\% & 0.0\% \\
Provincial expansion & All & 62378 & 93.7\% & 4.7\% & 1.4\% & 0.1\% & 0.0\% \\
\bottomrule
\end{tabular}%
}
\end{table}

\paragraph{Level and province breakdown.}
Table~\ref{tab:ai-level-province} disaggregates the same predictions by the
administrative level of the later document and by province. The directional
share falls steeply down the administrative ladder (national 9.7\%,
ministerial 7.6\%, provincial 1.6\%), and provinces with at least 300 pairs
span a thirty-fold range, from Anhui (15.5\%) and Beijing (8.4\%) down to
Heilongjiang (0.6\%) and Jilin (0.5\%). Mean ensemble confidence is
comparable across frames (0.968 central vs.\ 0.975 provincial), so the
gradient is not a confidence artifact.

\begin{table}[t]
\centering
\footnotesize
\caption{AI-policy directional change by administrative level of the later document (top) and by province (bottom; provinces with $n \geq 300$ pairs). Directional = expansion + restriction + reversal; remaining pairs are consistency or irrelevant.}
\label{tab:ai-level-province}
\begin{tabular}{lrrrrr}
\toprule
Level of $B$ & $n$ & Exp. & Res. & Rev. & Directional \\
\midrule
National & 597 & 9.5\% & 0.2\% & 0.0\% & 9.7\% \\
Ministerial & 3078 & 7.2\% & 0.2\% & 0.1\% & 7.6\% \\
Provincial & 62378 & 1.4\% & 0.1\% & 0.0\% & 1.6\% \\
\midrule
Province of $B$ & $n$ & Exp. & Res. & Rev. & Directional \\
\midrule
Anhui & 477 & 14.7\% & 0.8\% & 0.0\% & 15.5\% \\
Beijing & 801 & 7.7\% & 0.6\% & 0.0\% & 8.4\% \\
Chongqing & 979 & 3.5\% & 0.0\% & 0.0\% & 3.5\% \\
Zhejiang & 1357 & 2.7\% & 0.0\% & 0.1\% & 2.9\% \\
Shanghai & 6437 & 2.1\% & 0.2\% & 0.1\% & 2.4\% \\
Ningxia & 666 & 1.8\% & 0.3\% & 0.0\% & 2.1\% \\
Jiangsu & 1918 & 2.0\% & 0.1\% & 0.0\% & 2.1\% \\
Hunan & 792 & 0.8\% & 0.8\% & 0.1\% & 1.6\% \\
Shandong & 1416 & 1.4\% & 0.1\% & 0.0\% & 1.5\% \\
Inner Mongolia & 3010 & 0.7\% & 0.5\% & 0.0\% & 1.2\% \\
Henan & 1374 & 1.2\% & 0.0\% & 0.0\% & 1.2\% \\
Tianjin & 371 & 0.8\% & 0.3\% & 0.0\% & 1.1\% \\
Heilongjiang & 6608 & 0.6\% & 0.0\% & 0.0\% & 0.6\% \\
Jilin & 17112 & 0.4\% & 0.1\% & 0.0\% & 0.5\% \\
\bottomrule
\end{tabular}
\end{table}

\section{Domain Match versus Direction Discrimination}\label{app:domain-vs-direction}
Our primary metric collapses the two-axis outputs into the five-category
relation label. A fair question follows: does a model trained and scored on
the collapsed label perform the \emph{domain-match} judgment (do the
directives govern the same policy domain?) as reliably as the
\emph{direction} judgment (what does the later directive do to the earlier
policy?)? We answer without any new model runs, by decomposing existing
five-way predictions back into the two axes: a prediction of any of the four
change categories counts as ``matched'' for the domain axis, and direction is
scored on the four-way judgment over the 729 gold-related pairs with
\emph{irrelevant} predictions counted as errors.

\begin{table}[t]
\centering
\footnotesize
\caption{Post-hoc decomposition of five-way predictions into the two axes. \emph{Domain match} is the binary judgment that the two directives govern the same policy domain (the binary axis), scored on all 1{,}000 pairs; \emph{direction} is the four-way judgment on the 729 gold-related pairs, with \emph{irrelevant} or unparsed predictions counted as errors. \emph{Rec.} is the share of gold-related pairs the system keeps in the related set; \emph{cond.\ acc} is direction accuracy restricted to those pairs.}
\label{tab:domain-vs-direction}
\setlength{\tabcolsep}{3pt}
\begin{tabular}{lccc cccc c}
\toprule
& \multicolumn{3}{c}{domain match (binary)} & \multicolumn{4}{c}{direction (4-way)} & \\
\cmidrule(lr){2-4}\cmidrule(lr){5-8}
System & acc & $\kappa$ & F1 & m-F1 & $\kappa$ & acc & cond.\ acc & rec.\\
\midrule
Qwen3-14B (zero-shot) & 0.751 & 0.161 & 0.851 & 0.330 & 0.248 & 0.479 & 0.490 & 97.7\% \\
Qwen3-32B (zero-shot) & 0.758 & 0.185 & 0.855 & 0.451 & 0.444 & 0.599 & 0.611 & 98.1\% \\
GPT-5 (zero-shot) & 0.673 & 0.262 & 0.758 & 0.345 & 0.216 & 0.353 & 0.501 & 70.4\% \\
Qwen3-14B LoRA SFT (5-fold pooled) & 0.774 & 0.431 & 0.845 & 0.556 & 0.536 & 0.639 & 0.759 & 84.2\% \\
\bottomrule
\end{tabular}
\end{table}

Table~\ref{tab:domain-vs-direction} shows three things. First, zero-shot
models fail the domain boundary in family-specific ways: the Qwen models are
over-inclusive (keeping $\sim$98\% of gold-related pairs as related, which
collapses their binary domain-match agreement to $\kappa \approx 0.16$--$0.19$
by flooding the unrelated class), while GPT-5 is over-strict (recalling only
70.4\% of gold-related pairs). Second, once a pair is correctly held in the
related set, zero-shot direction accuracy is markedly better (0.49--0.61) than
the systems' headline five-way scores suggest --- the boundary, not the
direction, is where zero-shot failure concentrates. Third, fine-tuning
repairs both axes but leaves the boundary the harder one: the fine-tuned
model's binary domain-match agreement ($\kappa = 0.431$) remains well below
the human boundary ceiling ($\kappa = 0.730$), while its conditional
direction accuracy reaches 0.759. This mirrors the human data, where the
relevance boundary --- not graded direction --- was the main source of
inter-coder divergence, and it explains why adjudication efforts
(Appendix~\ref{app:adjudication}) concentrate on that boundary.

\subsection{Error Rates by Similarity, Separately and Jointly}
Table~\ref{tab:error-bands} breaks both axes' error rates down by document
similarity, and Table~\ref{tab:error-taxonomy} gives the joint taxonomy over
all pairs. Three patterns stand out. First, zero-shot domain-match errors peak
where similarity misleads, in family-specific directions: the over-inclusive
Qwen model errs most in the mid-similarity lure zone (0.39 in the 0.6--0.7
band), while over-strict GPT-5 errs most on high-similarity paraphrase
reissues (0.39 in the 0.8--0.9 band). Second, on the lowest-similarity
pairs---the cross-domain reissues of the kind Figure~1 of the main paper
illustrates---zero-shot direction error reaches 0.73--0.77 even when the
domain-match judgment is correct. Third, fine-tuning flattens the domain-error
curve across bands (0.19--0.23) and halves conditional direction error
(0.14--0.32), but the boundary share of its remaining errors rises to 60\%:
the better a model gets at direction, the more its residual errors are
boundary errors. This is the quantitative case for scoring the two axes
separately.

\begin{table}[t]
\centering
\footnotesize
\caption{Per-axis error rates by BGE cosine-similarity band. \emph{Domain} = wrong binary domain-match judgment; \emph{dir.} = wrong direction on gold-related pairs (boundary misses count as errors); \emph{dir.\ $|$ matched} = wrong direction given the pair was correctly kept as related.}
\label{tab:error-bands}
\setlength{\tabcolsep}{3.5pt}
\begin{tabular}{llrccc}
\toprule
System & band & $n$ & domain & dir. & dir.\ $|$ matched \\
\midrule
Qwen3-14B zero-shot & [0.0, 0.6) & 19 & 0.158 & 0.769 & 0.727 \\
 & [0.6, 0.7) & 151 & 0.391 & 0.520 & 0.429 \\
 & [0.7, 0.8) & 89 & 0.315 & 0.633 & 0.633 \\
 & [0.8, 0.9) & 427 & 0.173 & 0.390 & 0.390 \\
 & [0.9, 1.0) & 186 & 0.226 & 0.639 & 0.634 \\
\midrule
GPT-5 zero-shot & [0.0, 0.6) & 19 & 0.158 & 0.462 & 0.300 \\
 & [0.6, 0.7) & 151 & 0.245 & 0.680 & 0.400 \\
 & [0.7, 0.8) & 89 & 0.258 & 0.467 & 0.333 \\
 & [0.8, 0.9) & 427 & 0.389 & 0.650 & 0.438 \\
 & [0.9, 1.0) & 186 & 0.333 & 0.722 & 0.643 \\
\midrule
Qwen3-14B LoRA SFT & [0.0, 0.6) & 19 & 0.316 & 0.538 & 0.143 \\
 & [0.6, 0.7) & 151 & 0.225 & 0.560 & 0.250 \\
 & [0.7, 0.8) & 89 & 0.225 & 0.383 & 0.275 \\
 & [0.8, 0.9) & 427 & 0.222 & 0.305 & 0.203 \\
 & [0.9, 1.0) & 186 & 0.194 & 0.389 & 0.323 \\
\bottomrule
\end{tabular}
\end{table}

\begin{table}[t]
\centering
\footnotesize
\caption{Joint error taxonomy over all 1{,}000 pairs. \emph{Pure direction}: domain judgment correct, collapsed label wrong. \emph{Boundary-involved}: domain judgment wrong (the collapsed label is then almost necessarily wrong too). Shares are of all pairs, with the boundary share of all errors in the last column.}
\label{tab:error-taxonomy}
\begin{tabular}{lcccc}
\toprule
System & total err. & pure dir. & boundary & \% of errors \\
\midrule
Qwen3-14B zero-shot & 0.612 & 0.363 & 0.249 & 41\% \\
GPT-5 zero-shot & 0.583 & 0.256 & 0.327 & 56\% \\
Qwen3-14B LoRA SFT & 0.374 & 0.148 & 0.226 & 60\% \\
\bottomrule
\end{tabular}
\end{table}

\section{Formulation Variants: Joint, Cascade, and Direct}\label{app:twoaxis-note}
The dual-axis outputs can be elicited three ways: \emph{jointly} (both outputs
in one pass), in \emph{cascade} (domain match first, direction conditional on
a match), or \emph{directly} (the collapsed five-category judgment in one
pass, absorbing the domain axis). Table~\ref{tab:formulation-variants}
re-scores all three under the collapsed five-category metric for three
open-weight models. The three elicitations are predictively comparable---no
variant dominates---which is why the main paper scores the collapsed relation
as the primary metric. The oracle rows give the ceiling from a perfect domain-match
judgment: roughly +0.17 macro-F1 across model sizes, the quantitative
motivation for keeping the two axes separate in analysis even though the
elicitation format itself does not matter.

\begin{table}[t]
\centering
\footnotesize
\caption{Elicitation variants of the dual-axis task, scored on the collapsed five-category metric. \emph{Joint} predicts both outputs in one pass; \emph{cascade} predicts the domain match first and the direction second; \emph{direct} predicts only the collapsed relation; \emph{oracle} replaces the domain judgment with gold.}
\label{tab:formulation-variants}
\begin{tabular}{llccc}
\toprule
Model & elicitation & macro-F1 & $\kappa$ & acc \\
\midrule
qwen3-32b & joint (two outputs, one pass) & 0.447 & 0.333 & 0.477 \\
 & cascade (domain gate, then direction) & 0.428 & 0.303 & 0.442 \\
 & direct (collapsed five-category) & 0.462 & 0.322 & 0.474 \\
 & oracle domain gate (ceiling) & 0.618 & 0.576 & 0.666 \\
\midrule
qwen3-4b & joint (two outputs, one pass) & 0.238 & 0.107 & 0.324 \\
 & cascade (domain gate, then direction) & 0.406 & 0.265 & 0.411 \\
 & direct (collapsed five-category) & 0.361 & 0.250 & 0.401 \\
 & oracle domain gate (ceiling) & 0.619 & 0.559 & 0.653 \\
\midrule
qwen3-8b & joint (two outputs, one pass) & 0.355 & 0.219 & 0.388 \\
 & cascade (domain gate, then direction) & 0.322 & 0.194 & 0.347 \\
 & direct (collapsed five-category) & 0.391 & 0.266 & 0.422 \\
 & oracle domain gate (ceiling) & 0.519 & 0.470 & 0.578 \\
\bottomrule
\end{tabular}
\end{table}

\paragraph{Frontier flagships under the joint prompt.}
Table~\ref{tab:dualaxis_api} re-runs the four frontier flagships under the
original joint two-output prompt, scoring the binary domain axis and the
collapsed relation separately. Collapsed performance is within noise of the
direct prompt for GPT-5 and modestly lower for the others, with the ranking
preserved --- the elicitation format is not the bottleneck. The domain axis,
however, exposes the effect of the original prompt's erroneous
``$\sim$90\% related'' prior (measured rate: 72.9\%): all four models declare
95\%+ of pairs related, and binary domain-match agreement compresses to
$\kappa = 0.13$--$0.17$. All $\sim$4{,}000 outputs satisfied the
$d=0 \Leftrightarrow$ \textsc{unrelated} constraint exactly.
A corrected-prior variant of the joint prompt (73\% related) repairs part of
the domain-axis distortion (binary $\kappa$ rises on all four models, by
$+0.001$ to $+0.060$) while the collapsed metric stays flat or improves
slightly ($\Delta$macro-F1 $-0.001$ to $+0.024$; GPT-5's corrected dual-axis
run is its best overall, 0.440) ---
confirming that the prior bug distorted the domain axis specifically.

\begin{table*}
\centering
\footnotesize
\setlength{\tabcolsep}{4pt}
\caption{Frontier API flagships under the dual-axis (two-output) formulation: joint same-category and logical-relationship prediction on 1,000 gold pairs, collapsed to the five-way scheme for scoring. Each model appears under the original prompt and under a variant with the corrected related-pair prior (73\%).}
\label{tab:dualaxis_api}
\resizebox{\textwidth}{!}{%
\begin{tabular}{lrrrrrrrrrrrrr}
\toprule
slug & coverage & same\_acc & same\_kappa & same\_F1 & macroF1 & kappa & acc & F1\_irrelevant & F1\_consistency & F1\_restriction & F1\_expansion & F1\_reversal & constraint\_viol\_rate \\
\midrule
claude\_sonnet\_45\_\_dualaxis\_\_t1500 & 0.999 & 0.755 & 0.170 & 0.854 & 0.332 & 0.216 & 0.397 & 0.242 & 0.518 & 0.180 & 0.410 & 0.312 & 0.000 \\
claude\_sonnet\_45\_\_dualaxis\_cp\_\_t1500 & 0.999 & 0.757 & 0.184 & 0.855 & 0.331 & 0.213 & 0.396 & 0.262 & 0.512 & 0.152 & 0.417 & 0.312 & 0.000 \\
deepseek\_chat\_\_dualaxis\_\_t1500 & 1.000 & 0.751 & 0.142 & 0.852 & 0.362 & 0.259 & 0.424 & 0.204 & 0.550 & 0.197 & 0.474 & 0.385 & 0.000 \\
deepseek\_chat\_\_dualaxis\_cp\_\_t1500 & 1.000 & 0.759 & 0.202 & 0.855 & 0.368 & 0.266 & 0.429 & 0.294 & 0.552 & 0.183 & 0.468 & 0.344 & 0.001 \\
gemini\_25\_pro\_\_dualaxis\_\_t1500 & 1.000 & 0.755 & 0.161 & 0.854 & 0.388 & 0.249 & 0.418 & 0.227 & 0.520 & 0.274 & 0.365 & 0.554 & 0.000 \\
gemini\_25\_pro\_\_dualaxis\_cp\_\_t1500 & 1.000 & 0.754 & 0.162 & 0.854 & 0.402 & 0.266 & 0.431 & 0.231 & 0.527 & 0.294 & 0.403 & 0.554 & 0.000 \\
gpt\_5\_\_dualaxis\_\_t1500 & 1.000 & 0.750 & 0.135 & 0.852 & 0.416 & 0.279 & 0.433 & 0.194 & 0.545 & 0.385 & 0.374 & 0.584 & 0.000 \\
gpt\_5\_\_dualaxis\_cp\_\_t1500 & 1.000 & 0.754 & 0.170 & 0.853 & 0.440 & 0.307 & 0.456 & 0.245 & 0.568 & 0.420 & 0.390 & 0.577 & 0.000 \\
\bottomrule
\end{tabular}%
}
\end{table*}

\section{Training Regime, Sample Size, and Model-Family Bias}\label{app:training-bias}

\paragraph{Zero-shot, few-shot, and fine-tuned regimes.}
Table~\ref{tab:training-regime} compares the three regimes directly. Few-shot
prompting improves the frontier models only modestly, whereas expert-assisted
fine-tuning produces the large gain. This comparison is the basis for the main
paper's claim that the task is not solved by prompt engineering alone. The
silver row uses model-generated labels and is not additional expert annotation.

\begin{table*}[t]
\centering
\footnotesize
\setlength{\tabcolsep}{5pt}
\caption{Training-regime comparison on the expert gold set. Zero-shot rows use the main prompt for Qwen3-32B and the direct five-category prompt at 1,500 characters for frontier models; few-shot rows use the class-balanced corrected-prior prompt.}
\label{tab:training-regime}
\begin{tabular}{llrrrrr}
\toprule
Regime & Model & $n$ & coverage & macro-F1 & $\kappa$ & acc. \\
\midrule
Zero-shot & Qwen3-32B & 1000 & 1.000 & 0.452 & 0.337 & 0.480 \\
Zero-shot & GPT-5 & 1000 & 1.000 & 0.420 & 0.256 & 0.417 \\
Zero-shot & Gemini 2.5 Pro & 1000 & 0.998 & 0.418 & 0.273 & 0.436 \\
Zero-shot & Claude Sonnet 4.5 & 1000 & 0.999 & 0.412 & 0.265 & 0.432 \\
Zero-shot & DeepSeek-V3 & 1000 & 1.000 & 0.406 & 0.282 & 0.440 \\
Few-shot & GPT-5 & 995 & 0.999 & 0.445 & 0.279 & 0.436 \\
Few-shot & Gemini 2.5 Pro & 995 & 0.998 & 0.419 & 0.249 & 0.418 \\
Few-shot & Claude Sonnet 4.5 & 995 & 0.999 & 0.433 & 0.259 & 0.430 \\
Fine-tuned & Qwen3-14B LoRA & 1000 & 1.000 & 0.634 & 0.524 & 0.626 \\
Fine-tuned + model silver & Qwen3-14B LoRA & 1000 & 1.000 & 0.638 & 0.529 & 0.631 \\
\bottomrule
\end{tabular}
\end{table*}

\IfFileExists{generated/learning_curve.complete}{%
\paragraph{Annotation sample size.}
To estimate how much expert annotation is enough, we train the same Qwen3-14B
LoRA recipe on nested, label- and era-stratified subsets of each training fold
($n=50,100,200,400$) and compare them with the full fold training data and the
zero-shot Qwen3-14B control (Table~\ref{tab:learning-curve}). The curve is
evaluated on the unchanged held-out folds, so sample size is the only intended
manipulation. Performance rises monotonically with expert sample size, with
the largest absolute gain between zero-shot and 50 pairs and continued gains
through the full 804-pair fold.

\begin{table}[t]
\centering
\footnotesize
\caption{Expert-annotation learning curve for the five-way classifier under the unchanged LoRA recipe of the main paper.}
\label{tab:learning-curve}
\begin{tabular}{lrlccc}
\toprule
Setting & $n_\mathrm{train}$ & cov\% & macro-F1 & $\kappa$ & acc \\
\midrule
Qwen3-14B zero-shot (main prompt) & 0 & 100.0 & 0.342 & 0.202 & 0.388 \\
Qwen3-14B LoRA SFT (n=50) & 50 & 100.0 & 0.448 & 0.315 & 0.473 \\
Qwen3-14B LoRA SFT (n=100) & 100 & 100.0 & 0.510 & 0.361 & 0.498 \\
Qwen3-14B LoRA SFT (n=200) & 200 & 100.0 & 0.550 & 0.417 & 0.543 \\
Qwen3-14B LoRA SFT (n=400) & 400 & 100.0 & 0.587 & 0.460 & 0.575 \\
Qwen3-14B LoRA SFT (full gold) & 804 & 100.0 & 0.634 & 0.524 & 0.626 \\
\bottomrule
\end{tabular}
\end{table}

\IfFileExists{generated/fig_learning_curve.pdf}{%
\begin{figure}[t]
\centering
\includegraphics[width=0.9\columnwidth]{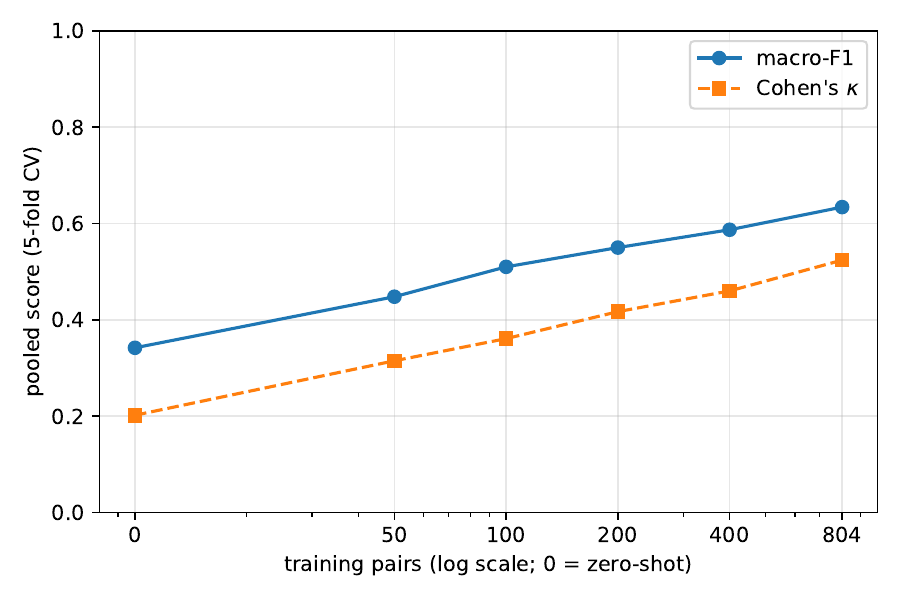}
\caption{Five-way classification performance by expert-annotation sample size.}
\label{fig:learning-curve}
\end{figure}}{}
}{%
}

\paragraph{Rare-category reliability.}
Expansion, restriction, and reversal have the
smallest expert-labeled supports. Table~\ref{tab:rare-class-bootstrap}
therefore reports pair-level bootstrap confidence intervals for per-class F1,
rather than only point estimates. The intervals make clear where the benchmark
is statistically fragile and where fine-tuning gains are large relative to
sampling uncertainty.

\begin{table*}[t]
\centering
\footnotesize
\caption{Per-class F1 with 95\% bootstrap confidence intervals for the best
open-weight zero-shot model, a frontier flagship, and the fine-tuned model.}
\label{tab:rare-class-bootstrap}
\begin{tabular}{l r rrr}
\toprule
 & & \multicolumn{3}{c}{F1 [95\% CI]} \\
\cmidrule(lr){3-5}
Class & $N_{gold}$ & Qwen3-32B (zs) & GPT-5 (zs) & Qwen3-14B (SFT) \\
\midrule
CONS & 255 & 0.567 [0.522,0.610] & 0.199 [0.145,0.253] & 0.710 [0.664,0.753] \\
EXPAN & 171 & 0.531 [0.476,0.585] & 0.433 [0.378,0.486] & 0.551 [0.487,0.612] \\
RESTR & 176 & 0.425 [0.350,0.495] & 0.357 [0.289,0.428] & 0.544 [0.481,0.601] \\
CONTR & 127 & 0.477 [0.375,0.568] & 0.618 [0.537,0.690] & 0.780 [0.716,0.833] \\
UNREL & 271 & 0.262 [0.199,0.327] & 0.495 [0.448,0.542] & 0.586 [0.536,0.633] \\
\bottomrule
\end{tabular}

\end{table*}

\begin{figure}[t]
\centering
\includegraphics[width=0.9\columnwidth]{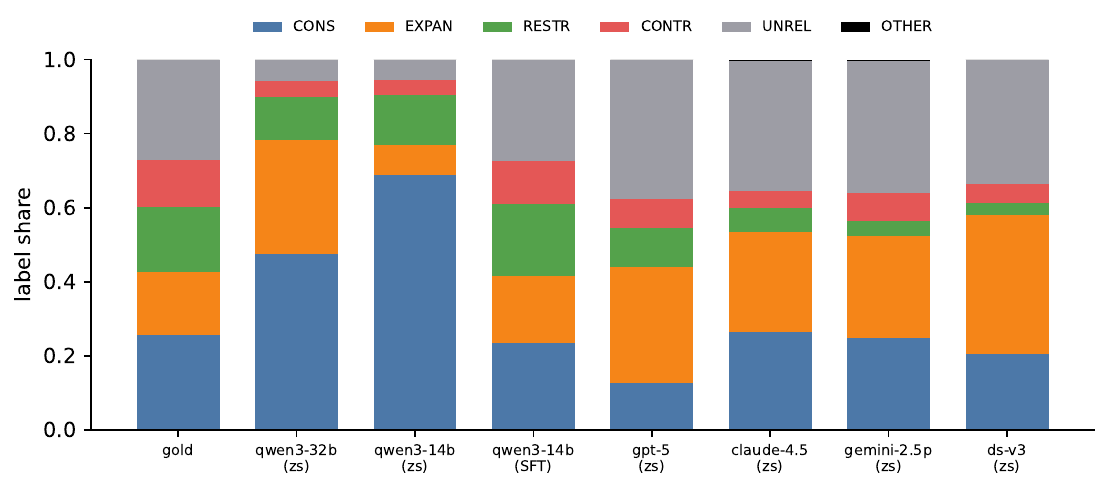}
\caption{Predicted-label distribution shift from gold by model family. Lower
TVD means the model's predicted label marginal is closer to the expert gold
marginal.}
\label{fig:label-shift}
\end{figure}

\paragraph{Model-family label bias.}
We next test whether model families systematically impose different label
priors on Chinese policy texts. For every model we compare its predicted
five-class distribution with the gold distribution on the same 1,000 pairs,
using total variation distance (TVD) and KL divergence; we also record which
class is most over- or under-predicted. Figure~\ref{fig:label-shift} shows that
frontier models are less skewed than small open models, but still exhibit
systematic expansion/restriction and consistency/irrelevance biases.
Table~\ref{tab:model-family-bias} gives the per-model values.

\begin{table*}[t]
\centering
\footnotesize
\setlength{\tabcolsep}{3pt}
\caption{Label-distribution bias by model and family on the same 1,000-pair
gold set. `Over' and `Under' name the classes with the largest positive and
negative predicted-minus-gold marginal shifts.}
\label{tab:model-family-bias}
\resizebox{\textwidth}{!}{%
\begin{tabular}{lllrrrrll}
\toprule
Model & Family & Cond. & Cov. & TVD & KL & MacroRec. & Over & Under \\
\midrule
claude-sonnet-4.5 & Claude & zs & 0.999 & 0.193 & 0.116 & 0.411 & EXPAN & RESTR \\
deepseek-r1-distill-qwen-14b & DeepSeek & zs & 1.000 & 0.526 & 0.612 & 0.317 & CONS & UNREL \\
deepseek-v3 & DeepSeek & zs & 1.000 & 0.450 & 0.487 & 0.362 & CONS & UNREL \\
deepseek-v3-chat & DeepSeek & zs & 1.000 & 0.268 & 0.216 & 0.428 & EXPAN & RESTR \\
glm-4-9b-chat & GLM & zs & 0.822 & 0.437 & 1.082 & 0.303 & EXPAN & RESTR \\
gpt-5 & GPT & zs & 1.000 & 0.248 & 0.132 & 0.428 & EXPAN & CONS \\
gpt-5-mini & GPT & zs & 1.000 & 0.391 & 0.428 & 0.438 & EXPAN & UNREL \\
gemini-2.5-flash & Gemini & zs & 1.000 & 0.359 & 0.301 & 0.449 & CONS & UNREL \\
gemini-2.5-pro & Gemini & zs & 0.998 & 0.195 & 0.130 & 0.425 & EXPAN & RESTR \\
gemma-2-9b-it & Gemma & zs & 1.000 & 0.403 & 0.396 & 0.316 & CONS & UNREL \\
gemma-7b-it & Gemma & zs & 0.984 & 0.390 & 0.419 & 0.242 & CONS & RESTR \\
internlm2\_5-7b-chat & InternLM & zs & 0.993 & 0.469 & 0.612 & 0.263 & CONS & RESTR \\
llama-3.1-8b-instruct & Llama & zs & 1.000 & 0.723 & 1.222 & 0.217 & CONS & UNREL \\
llama-3.3-70b & Llama & zs & 0.999 & 0.434 & 0.430 & 0.386 & CONS & UNREL \\
mimo-7b-rl & MiMo & zs & 0.915 & 0.364 & 0.561 & 0.321 & EXPAN & UNREL \\
mistral-7b-instruct-v0.2 & Mistral & zs & 0.257 & 0.743 & 4.741 & 0.083 & CONS & UNREL \\
qwen2.5-14b-instruct & Qwen & zs & 0.984 & 0.392 & 0.451 & 0.395 & EXPAN & RESTR \\
qwen2.5-7b-instruct & Qwen & zs & 1.000 & 0.472 & 0.549 & 0.277 & CONS & RESTR \\
qwen3-14b & Qwen & SFT & 1.000 & 0.033 & 0.003 & 0.631 & RESTR & CONS \\
qwen3-14b & Qwen & zs & 1.000 & 0.434 & 0.448 & 0.358 & CONS & UNREL \\
qwen3-32b & Qwen & zs & 1.000 & 0.356 & 0.288 & 0.478 & CONS & UNREL \\
qwen3-4b & Qwen & zs & 1.000 & 0.575 & 0.809 & 0.289 & CONS & UNREL \\
qwen3-8b & Qwen & zs & 1.000 & 0.376 & 0.338 & 0.378 & CONS & UNREL \\
\bottomrule
\end{tabular}

}
\end{table*}

\paragraph{Does expert fine-tuning correct the bias?}
For Qwen3-14B, we can answer this directly because the same base model has
zero-shot predictions and five-fold SFT predictions on the same pairs.
Table~\ref{tab:qwen-correction} shows that fine-tuning sharply reduces the
label-marginal shift and repairs the zero-shot model's recall collapse on
irrelevant and reversal pairs.

\begin{table}[t]
\centering
\footnotesize
\caption{Qwen3-14B before and after expert fine-tuning on the same 1,000 pairs.}
\label{tab:qwen-correction}
\begin{tabular}{lrrrrrrr}
\toprule
Class & $N_{gold}$ & \multicolumn{2}{c}{Precision} & \multicolumn{2}{c}{Recall} & \multicolumn{2}{c}{F1} \\
\cmidrule(lr){3-4}\cmidrule(lr){5-6}\cmidrule(lr){7-8}
 & & ZS & SFT & ZS & SFT & ZS & SFT \\
\midrule
CONS & 255 & 0.340 & 0.740 & 0.918 & 0.682 & 0.496 & 0.710 \\
EXPAN & 171 & 0.469 & 0.536 & 0.222 & 0.567 & 0.302 & 0.551 \\
RESTR & 176 & 0.336 & 0.518 & 0.256 & 0.574 & 0.290 & 0.544 \\
CONTR & 127 & 0.800 & 0.825 & 0.252 & 0.740 & 0.383 & 0.780 \\
UNREL & 271 & 0.696 & 0.582 & 0.144 & 0.590 & 0.239 & 0.586 \\
\bottomrule
\end{tabular}

\end{table}

\IfFileExists{generated/tab_family_correction.tex}{%
\paragraph{Cross-family correction.}
A matched non-Chinese-base Mistral arm tests whether the correction observed
for Qwen generalizes across model families. Table~\ref{tab:family-correction}
shows that the same expert-labeled training data produce a large gain in
Mistral-7B as well, although the fine-tuned Mistral remains below fine-tuned
Qwen3-14B. Figure~\ref{fig:family-correction} summarizes the two metrics.
\begin{table}[t]
\centering
\footnotesize
\caption{Cross-family correction from expert fine-tuning. Mistral-7B-Instruct-v0.2 and Qwen3-14B use the same five folds and LoRA recipe; lower-coverage zero-shot Mistral is scored under the zero-shot convention that counts unparsed outputs as errors, so coverage is reported alongside performance.}
\label{tab:family-correction}
\resizebox{\textwidth}{!}{%
\begin{tabular}{l l r r r r r r r r r}
\toprule
\textbf{Family} & \textbf{Condition} & \textbf{Cov\%} & \textbf{Macro-F1} & \textbf{$\kappa$} & \textbf{Acc} & \textbf{cons} & \textbf{expa} & \textbf{rest} & \textbf{reve} & \textbf{irre} \\
\midrule
Mistral-7B-Instruct-v0.2 & zero-shot & 84.2 & 0.180 & 0.182 & 0.226 & 0.373 & 0.043 & 0.029 & 0.000 & 0.457 \\
 & SFT (5-fold LoRA) & 100.0 & 0.608 & 0.492 & 0.600 & 0.690 & 0.523 & 0.516 & 0.770 & 0.539 \\
\midrule
Qwen3-14B & zero-shot & 100.0 & 0.342 & 0.202 & 0.388 & 0.496 & 0.302 & 0.290 & 0.383 & 0.239 \\
 & SFT (5-fold LoRA) & 100.0 & 0.634 & 0.524 & 0.626 & 0.710 & 0.551 & 0.544 & 0.780 & 0.586 \\
\midrule
 & SFT - zero-shot & 15.8 & 0.428 & 0.310 & 0.374 & 0.317 & 0.480 & 0.487 & 0.770 & 0.082 \\
 & SFT - zero-shot & 0.0 & 0.292 & 0.322 & 0.238 & 0.214 & 0.249 & 0.254 & 0.397 & 0.347 \\
\bottomrule
\end{tabular}%
}
\end{table}

\begin{figure}[t]
\centering
\includegraphics[width=0.9\columnwidth]{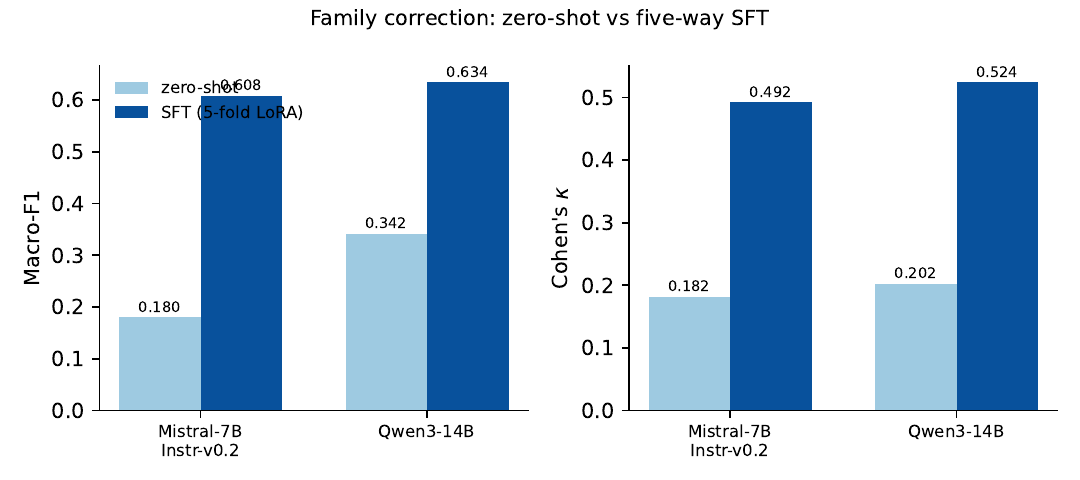}
\caption{Zero-shot versus fine-tuned performance for Mistral-7B-Instruct-v0.2
and Qwen3-14B under the same five-fold recipe.}
\label{fig:family-correction}
\end{figure}
}{%
}

\section{Additional Training and Test-Time Techniques}\label{app:techniques}
Beyond the main-paper recipe (LoRA SFT of Qwen3-14B), we compare a sequence
of training and test-time techniques on the same fold-0 validation split:
GRPO (reinforcement learning on the five-way reward), DPO and KTO
(preference-style objectives), RAFT (rejection-sampling SFT), and
self-consistency (majority vote over $k{=}8$ samples at test time).
Table~\ref{tab:wavee} shows that self-consistency is the only arm that
improves both macro-F1 and $\kappa$ over the fold-0 SFT baseline; DPO is
effectively flat, while RAFT and GRPO fall below baseline. KTO produced no
score because the TRL 0.23.1 trainer failed deterministically after 79 of 90
steps on two different GPU nodes. The canonical 80/20 SFT row is reported
separately because it uses a different held-out split.

\begin{table*}[t]
\centering
\footnotesize
\setlength{\tabcolsep}{5pt}
\caption{Training- and test-time technique comparison. Fold-0 validation arms use $n=196$; the canonical 80/20 row uses its held-out $n=204$ split and is reported separately from the fold-0 comparison. KTO is included for completeness but produced no score because TRL 0.23.1 failed deterministically at training step 79/90 across two GPU nodes.}
\label{tab:wavee}
\begin{tabular}{lrrrrl}
\toprule
Technique & $n$ & macro-F1 & $\kappa$ & acc. & note \\
\midrule
SFT fold-0 baseline & 196 & 0.635 & 0.533 & 0.633 & replication \\
Self-consistency $k=8$ & 196 & 0.638 & 0.553 & 0.648 & test-time vote \\
RAFT & 196 & 0.623 & 0.524 & 0.622 &  \\
DPO & 196 & 0.632 & 0.538 & 0.638 &  \\
GRPO & 196 & 0.622 & 0.519 & 0.617 &  \\
KTO & -- & -- & -- & -- & deterministic TRL failure \\
SFT canonical 80/20 & 204 & 0.621 & 0.519 & 0.623 &  \\
\bottomrule
\end{tabular}
\end{table*}

We also apply the five cross-validated adapters as a teacher ensemble to a
5,000-pair unlabeled slice. Table~\ref{tab:teacher-agreement} reports the
agreement and predicted-label distribution used as the distillation corpus
for follow-up training.

\begin{table}[t]
\centering
\footnotesize
\caption{Five-adapter teacher labeling of the 5,000-pair unlabeled slice.}
\label{tab:teacher-agreement}
\begin{tabular}{lr}
\toprule
Quantity & Value \\
\midrule
Pairs & 5,000 \\
Unanimous (5/5) & 3,522 (70.4\%) \\
4/5 agreement & 765 \\
3/5 agreement & 645 \\
2/5 agreement & 68 \\
Mean pairwise adapter agreement & 0.849 \\
\midrule
Irrelevant & 1,870 \\
Expansion & 1,537 \\
Restriction & 1,100 \\
Reversal & 251 \\
Consistency & 242 \\
\bottomrule
\end{tabular}
\end{table}

\section{Context-Length Robustness}\label{app:context}
Two complementary analyses support the main paper's claim that the
1,500-character truncation does not hide decisive evidence.

\paragraph{Evidence coverage.} On full texts, the decisive repeal cue
(cessation/repeal language) appears within the first 1,500 characters of the
later document in 99\% of reversal pairs (median position 268 characters,
90th percentile 1,038). Cross-references to the earlier document's exact
title occur in only 19\% of related pairs, nearly all within the window.

\paragraph{Context sweep.} Extending the input window to 4,000 and 8,000
characters, and to full text where budget allowed, leaves five-category
performance flat or slightly negative on GPT-5 and DeepSeek-V3.
Table~\ref{tab:waveb-variants} reports all variant arms---few-shot with a
corrected class prior, multi-seed (temperature 0.7), and the context sweep.

\begin{table*}
\centering
\scriptsize
\setlength{\tabcolsep}{4pt}
\caption{Frontier flagship variants: few-shot, multi-seed (temperature 0.7), and context-length arms under the direct five-category prompt.}
\label{tab:waveb-variants}
\resizebox{\textwidth}{!}{%
\begin{tabular}{lrrrrrrrrrr}
\toprule
slug & n\_scored & coverage & macroF1 & kappa & acc & F1\_irrelevant & F1\_consistency & F1\_restriction & F1\_expansion & F1\_reversal \\
\midrule
deepseek\_chat\_\_direct5\_\_t0 & 1000 & 1.000 & 0.390 & 0.256 & 0.416 & 0.506 & 0.284 & 0.164 & 0.470 & 0.525 \\
deepseek\_chat\_\_direct5\_\_t4000 & 1000 & 0.999 & 0.391 & 0.258 & 0.420 & 0.510 & 0.322 & 0.141 & 0.475 & 0.506 \\
deepseek\_chat\_\_direct5\_\_t8000 & 1000 & 0.996 & 0.397 & 0.266 & 0.425 & 0.511 & 0.352 & 0.160 & 0.470 & 0.494 \\
gemini\_25\_pro\_\_direct5\_\_t1500\_\_temp0.7\_\_seed1 & 1000 & 0.998 & 0.434 & 0.290 & 0.449 & 0.521 & 0.398 & 0.237 & 0.465 & 0.546 \\
gemini\_25\_pro\_\_direct5\_\_t1500\_\_temp0.7\_\_seed2 & 1000 & 0.997 & 0.431 & 0.289 & 0.448 & 0.521 & 0.386 & 0.221 & 0.480 & 0.546 \\
gpt\_5\_\_direct5\_\_t1500\_\_temp0.7\_\_seed1 & 1000 & 1.000 & 0.425 & 0.262 & 0.423 & 0.514 & 0.216 & 0.337 & 0.420 & 0.638 \\
gpt\_5\_\_direct5\_\_t1500\_\_temp0.7\_\_seed2 & 1000 & 1.000 & 0.421 & 0.263 & 0.423 & 0.501 & 0.220 & 0.319 & 0.454 & 0.612 \\
gpt\_5\_\_direct5\_\_t4000 & 1000 & 1.000 & 0.407 & 0.244 & 0.407 & 0.486 & 0.197 & 0.304 & 0.436 & 0.611 \\
gpt\_5\_\_direct5\_\_t8000 & 1000 & 1.000 & 0.414 & 0.254 & 0.415 & 0.490 & 0.223 & 0.318 & 0.437 & 0.605 \\
claude\_sonnet\_45\_\_direct5\_5shot\_\_t1500 & 995 & 0.999 & 0.433 & 0.259 & 0.430 & 0.515 & 0.360 & 0.331 & 0.388 & 0.571 \\
gemini\_25\_pro\_\_direct5\_5shot\_\_t1500 & 995 & 0.998 & 0.419 & 0.249 & 0.418 & 0.500 & 0.307 & 0.317 & 0.432 & 0.540 \\
gpt\_5\_\_direct5\_5shot\_\_t1500 & 995 & 0.999 & 0.445 & 0.279 & 0.436 & 0.504 & 0.197 & 0.420 & 0.398 & 0.706 \\
\bottomrule
\end{tabular}%
}
\end{table*}

\section{The Policy-Signal Axis as Auxiliary Evidence}\label{app:color-fusion}
The corpus carries a second, cheaper measurement layer: a paragraph-level
policy-signal coding (prohibiting, pressuring, authorizing, encouraging,
aspirational; aggregated per document over actionable paragraphs only) that
tracks the \emph{strength of obligations} a document imposes. In dual-axis
terms, this signal speaks only to the direction axis---a shift toward harder
obligations suggests expansion, a shift toward softer ones suggests
restriction---and says nothing about whether two directives govern the same
policy domain. It is therefore a natural auxiliary signal, and an earlier version
of this work found that a similar corpus-wide signal helps only when learned
jointly and hurts as a hard rule. We re-test that lesson with the production
signal on the present benchmark.

A coverage caveat comes first: most benchmark documents are short approval
notices, and only 19.2\% of gold pairs have enough actionable paragraphs in
both documents for a usable signal profile. All arms below are nevertheless
scored on the full frame, with uncovered pairs counting as errors for the
signal-only rule, so the numbers are comparable to the main tables.

\begin{table}[t]
\centering
\footnotesize
\caption{Auxiliary policy-signal arms on the held-out folds. \emph{Signal-only rule} maps the between-document shift in the share of obligation-hardening paragraphs (prohibiting + pressuring, within each document's actionable paragraphs) to a direction label with a threshold tuned on the training folds; it never sees the texts and cannot express reversal. \emph{Prompt fusion} appends one signal-profile line per document to the prompt of the fine-tuned classifier. \emph{Hard-rule override} forces the classifier's answer to expansion/restriction whenever the signal shift exceeds the same threshold (negative control). Direction metrics are four-way on gold-related pairs; abstentions count as errors.}
\label{tab:color-arms}
\setlength{\tabcolsep}{3pt}
\begin{tabular}{l ccc ccc}
\toprule
& \multicolumn{3}{c}{five-way (all pairs)} & \multicolumn{3}{c}{direction (related pairs)} \\
\cmidrule(lr){2-4}\cmidrule(lr){5-7}
System & m-F1 & $\kappa$ & acc & m-F1 & $\kappa$ & cond.\ acc \\
\midrule
Signal-only rule (obligations) & -- & -- & -- & 0.074 & 0.023 & 0.431 \\
Signal-only rule (flexibility) & -- & -- & -- & 0.074 & 0.017 & 0.382 \\
Prompt fusion (signal line in prompt) & 0.638 & 0.530 & 0.629 & 0.565 & 0.558 & 0.755 \\
Fine-tuned classifier (no signal) & 0.649 & 0.548 & 0.643 & 0.572 & 0.576 & 0.759 \\
Hard-rule override (neg. control) & 0.631 & 0.524 & 0.623 & 0.552 & 0.545 & 0.725 \\
\bottomrule
\end{tabular}
\end{table}

Table~\ref{tab:color-arms} delivers a clean negative result, three ways.
First, the signal-only rule is near chance at pair level (direction
$\kappa = 0.02$): paragraph-color aggregates are too coarse to replace
reading. Second, appending each document's signal profile to the fine-tuned
classifier's prompt does not help ($\kappa$ 0.530 vs.\ 0.548 without the
line)---the model already extracts stronger direction evidence from the text
than the aggregate carries. Third, forcing the answer to follow the signal
(the hard rule) degrades performance exactly as the earlier fusion
experiment predicted ($\kappa = 0.524$). The auxiliary axis is real but
weak, and the residual gains in this task live in expert labels for the
boundary, not in side signals.

\section{Prompt Listings}\label{app:prompts}
All prompts below are verbatim from the released code
(\texttt{run\_formulation.py}, \texttt{benchmark.py}, and the frontier-runner
\texttt{run\_waveb.py}); line breaks are added for print only. Every input
carries title (max 200 characters), agency (max 40), year, and up to 1{,}500
characters of content per directive. Unparsed or refused outputs count as
errors in every reported score; we also report structured-output coverage.

\subsection*{M.1 Direct collapsed five-category prompt (headline runs)}
System message:
\begin{verbatim}
You are a Chinese policy analyst. Classify the logical relationship of
later policy B to earlier policy A as exactly one of CONSISTENCY,
EXPANSION, RESTRICTION, CONTRADICTION, UNRELATED.
  CONSISTENCY    same topic, no material change between A and B
  EXPANSION      same topic, B broadens A
  RESTRICTION    same topic, B narrows A
  CONTRADICTION  same topic, B reverses A
  UNRELATED      different topics
Output ONLY: {"logical_relationship": "ONE_OF_THE_FIVE"} /no_think
\end{verbatim}
User template (all prompt variants):
\begin{verbatim}
POLICY A (year={a_year}, agency={a_agency}):
title: {a_title}
content:
{a_content}

POLICY B (year={b_year}, agency={b_agency}):
title: {b_title}
content:
{b_content}
\end{verbatim}
(The joint and cascade prompts below additionally include the L1/L3 topic
metadata fields.)

\subsection*{M.2 Joint dual-axis prompt (two outputs, one pass)}
System message:
\begin{verbatim}
You are a Chinese policy analyst. You will judge whether two policy
documents (A and B) address the SAME TOPIC. Be LIBERAL: two policies
count as same-category if they share ANY substantive topical link -
same program, same subject area, same regulatory regime, even from
different agencies, jurisdictions, or years. Only mark UNRELATED when
the two policies clearly address DIFFERENT subjects (e.g., one is about
shipping infrastructure, the other about K-12 education). Output ONLY a
JSON object - no prose, no <think> tags. /no_think
\end{verbatim}
The user template appends annotation guidance and the two-label output
convention:
\begin{verbatim}
Annotation guidance:
- Same L1 or L3 topic is STRONG evidence of same_category=1.
- Different agencies / jurisdictions / years are NOT reasons to mark
  UNRELATED.
- Mark UNRELATED only when the substantive topics are clearly different.

Two labels:
  1. same_category: 1 if A and B share a topical link; 0 if their
     subjects are clearly different.
  2. logical_relationship: one of CONSISTENCY / EXPANSION / RESTRICTION
     / CONTRADICTION / UNRELATED (definitions as in M.1).
\end{verbatim}
An earlier version of the joint prompt additionally asserted an expected
related share of ``$\sim$90\%''; this prior was wrong (the measured rate is
72.9\%) and was corrected in all later runs, including the five-shot
arm below. The dual-axis frontier arm (Appendix~\ref{app:twoaxis-note})
deliberately re-ran the verbatim original prompt once to measure the prior's
effect on the domain axis; a corrected-prior dual-axis arm uses the sentence
in M.4.

\subsection*{M.3 Cascade prompts (domain gate, then direction)}
Stage 1 (domain match):
\begin{verbatim}
You are a Chinese policy analyst. Decide whether policies A and B address
the SAME substantive policy topic. Be liberal: any shared
program/regime/subject counts.
Output ONLY: {"same_category": 0_or_1} /no_think
\end{verbatim}
Stage 2 (direction, only for pairs gated as matched):
\begin{verbatim}
You are a Chinese policy analyst. A and B address the SAME policy topic.
Classify how later B changes earlier A as exactly one of CONSISTENCY,
EXPANSION, RESTRICTION, CONTRADICTION.
  CONSISTENCY no material change    EXPANSION B broadens A
  RESTRICTION B narrows A           CONTRADICTION B reverses A
Output ONLY: {"logical_relationship": "ONE_OF_THE_FOUR"} /no_think
\end{verbatim}

\subsection*{M.4 Class-balanced five-shot variant}
The five-shot arm prepends five worked demonstrations to the direct prompt
(M.1), one gold pair per class, selected deterministically (first
lexicographic pair id per class outside the evaluation set) and rendered as
abbreviated titles with their labels; evaluation pairs used as demos are
excluded from scoring (995 scored pairs). The class-prior sentence is the
corrected one:
\begin{verbatim}
Roughly 73% of pairs in this dataset share a topic; do not assume a pair
is related by default.
\end{verbatim}

\subsection*{M.5 Output parsing and scoring}
Outputs are parsed by extracting the first JSON object and reading the
label key(s), with a case-insensitive substring fallback over the valid
label set; anything else (refusals, malformed JSON, out-of-set labels) is
counted as an error, so every system is scored on every pair. For dual-axis
outputs, the collapsed relation is \textsc{unrelated} exactly when
\texttt{same\_category} $=0$; constraint violations are reported separately.

\section{Data Statement}\label{app:data}
We will publicly release: the 1,000-pair five-category labels; the document-group-disjoint
cross-validation splits; all prompts; raw model outputs for every reported
run (open-weight, frontier API, fine-tuned); the adjudication packet
instruments; and the scoring scripts that generate every table in this paper
and supplement. The release preserves source-document references and
documents the sampling frame (embedding retrieval with rare-class
enrichment) and annotation biases (the two-protocol relevance boundary
documented in the main paper).

\end{document}